\documentclass[runningheads]{llncs}

\usepackage{eccv}

\usepackage{eccvabbrv}
\usepackage{comment}
\usepackage{graphics}
\usepackage{multirow}

\usepackage{graphicx}
\usepackage{booktabs}
\usepackage{siunitx}
\usepackage{wrapfig}
\usepackage{microtype}

\usepackage[accsupp]{axessibility}  

\definecolor{cvprblue}{rgb}{0.21,0.49,0.74}
\usepackage[pagebackref,breaklinks,colorlinks,citecolor=cvprblue]{hyperref}

\usepackage{orcidlink}

\usepackage{xparse}

\NewDocumentCommand{\mbf}{m g}{%
  \IfNoValueTF{#2}{%
    \mathbf{#1}%
  }{%
    \mathbf{#1}_{#2}%
  }%
}
\NewDocumentCommand{\tbf}{m g}{%
  \IfNoValueTF{#2}{%
    \textbf{#1}%
  }{%
    \textbf{#1}_{#2}%
  }%
}

\usepackage{booktabs}
\usepackage{balance}
\usepackage{multirow}
\usepackage{overpic}
\usepackage{cite}
\usepackage[font=small,labelfont=bf]{caption}
\usepackage{color}
\usepackage{pifont}
\makeatletter
\@namedef{ver@everyshi.sty}{}
\makeatother
\usepackage{tikz}
\usetikzlibrary{patterns}

\usepackage{colortbl}
\usepackage{pgfplots}
\pgfplotsset{compat=1.17}
\usepackage{tabularx}
\usepackage{arydshln}
\usepackage{amssymb}
\usepackage{pifont}

 \usepackage{amsmath}
\usepackage{wasysym}%
\usepackage{enumitem}
\usepackage{wrapfig}
\usepackage{xspace}
\usepackage{tabu}
\usepackage[super]{nth}
\usepackage{scalefnt}
\usepackage{dsfont}
\usepackage{graphicx,calc}
\usepackage{algorithm}
\usepackage{algpseudocode}

\definecolor{darkgreen}{RGB}{0,153,51}
\definecolor{linkgreen}{RGB}{52,130,48}
\definecolor{LightCyan}{rgb}{0.87,0.92,0.96}
\definecolor{m_green}{RGB}{233, 254, 187}
\definecolor{m_orange}{RGB}{255, 212, 121}
\definecolor{m_red}{RGB}{255, 190, 188}
\definecolor{m_violet}{RGB}{215, 131, 255}
\definecolor{m_blue}{RGB}{186, 234, 255}
\definecolor{m_brown}{RGB}{255,212,120}
\definecolor{m_lightgreen}{RGB}{212,251,122}
\definecolor{notetext}{rgb}{0.7,0,0}

\definecolor{model_pink}{RGB}{235, 106, 164}
\definecolor{model_orange}{RGB}{250, 194, 122}
\definecolor{model_green}{RGB}{164, 210, 162}
\definecolor{model_gray}{RGB}{120, 120, 120}
\definecolor{model_yellow}{RGB}{251, 231, 171}
\definecolor{model_purple}{RGB}{190, 146, 211}

\newcommand{\cdforw}{\overrightarrow{\text{CD}}}
\newcommand{\cdbakw}{\overleftarrow{\text{CD}}}

\usepackage{amssymb}
\usepackage{pifont}
\newcommand{\cmark}{\ding{51}}%
\newcommand{\xmark}{\ding{55}}%

\def\ie{\emph{i.e.}\@\xspace} 
\def\cf{\emph{c.f.}\@\xspace} 
 
\def\etal{\emph{et al.}\@\xspace}

\newcommand{\parag}[1]{\vspace{-2px} \vskip8pt \noindent \textbf{#1}}

\newcommand{\name}{P2P-Bridge}

\newcolumntype{Y}{>{\centering\arraybackslash}X}
\newcolumntype{Z}{>{\raggedleft\arraybackslash}X}

\usepackage{array}
\newcolumntype{P}[1]{>{\centering\arraybackslash}p{#1}}
\newcolumntype{M}[1]{>{\centering\arraybackslash}m{#1}}

\renewcommand{\vec}[1]{\mathbf{#1}}

\definecolor{darkblue}{RGB}{60, 82, 145}
\definecolor{kingblue}{RGB}{65, 105, 225}

\newlength\myheight
\newlength\mydepth
\settototalheight\myheight{Xygp}
\colorlet{colorFst}{Green!25}       
\colorlet{colorSnd}{SpringGreen!45} 
\colorlet{colorTrd}{Yellow!30}      
\newcommand{\st}{\cellcolor{colorFst}\bf}   
\newcommand{\nd}{\cellcolor{colorSnd}}      
\newcommand{\rd}{\cellcolor{colorTrd}}      

\ExplSyntaxOn
\NewDocumentEnvironment{places}{mm}
 {
  \setlength{\tabcolsep}{0pt} 
  \dim_set:Nn \l_places_width_dim
   {
    (#1-\ht\strutbox-\dp\strutbox-2pt)/(#2)
   }
  \begin{tabular}{r @{\hspace{2pt}} *{#2}{c}}
 }
 {
  \end{tabular}
 }

\NewDocumentCommand{\place}{mm}
 {
  \seq_set_from_clist:Nn \l_places_images_in_seq { #2 }
  \seq_set_map:NNn \l_places_images_out_seq \l_places_images_in_seq { \places_set_image:n {##1} }
  \seq_put_left:Nn \l_places_images_out_seq
   {
    \begin{tabular}{c}\rotatebox[origin=c]{90}{\strut#1}\end{tabular}
   }
  \seq_use:Nn \l_places_images_out_seq { & } \\ \addlinespace
 }

\dim_new:N \l_places_width_dim
\seq_new:N \l_places_images_in_seq
\seq_new:N \l_places_images_out_seq

\cs_new_protected:Nn \places_set_image:n
 {
  \makebox[\l_places_width_dim]
   {
    \begin{tabular}{c}
    \includegraphics[
      width=\l_places_width_dim,
      height=\l_places_width_dim,
      keepaspectratio,
    ]{#1}
    \end{tabular}
   }
 }

\ExplSyntaxOff

\begin{document}

\title{P2P-Bridge: Diffusion Bridges for 3D Point Cloud Denoising} 

\author{Mathias Vogel\inst{1}
\and
Keisuke Tateno\inst{2}
\and
Marc Pollefeys\inst{1,3}
\and
Federico Tombari\inst{2,4}
\and
Marie-Julie Rakotosaona\inst{2,*}
\and
Francis Engelmann\inst{1,2,*}
}

\authorrunning{M.~Vogel et al.}

\institute{
  \textsuperscript{1}ETH Zurich \quad
  \textsuperscript{2}Google \quad
  \textsuperscript{3}Microsoft \quad
  \textsuperscript{4}TU Munich \quad
  \textsuperscript{*}equal contribution
}

\maketitle

\begin{abstract}
In this work, we tackle the task of point cloud denoising through a novel framework that adapts Diffusion Schrödinger bridges to points clouds.
Unlike previous approaches that predict point-wise displacements from point features or learned noise distributions,
our method learns an optimal transport plan between paired point clouds. 
Experiments on object datasets like PU-Net and real-world datasets such as ScanNet++ and ARKitScenes show that \name{} achieves significant improvements over existing methods.
While our approach demonstrates strong results using only point coordinates, we also show that incorporating additional features, such as color information or point-wise DINOv2 features, further enhances the performance. 
Code and pretrained models are available at \href{https://p2p-bridge.github.io}{https://p2p-bridge.github.io}.
\end{abstract}
\begin{figure}[t]
\centering
\includegraphics[width=\textwidth]{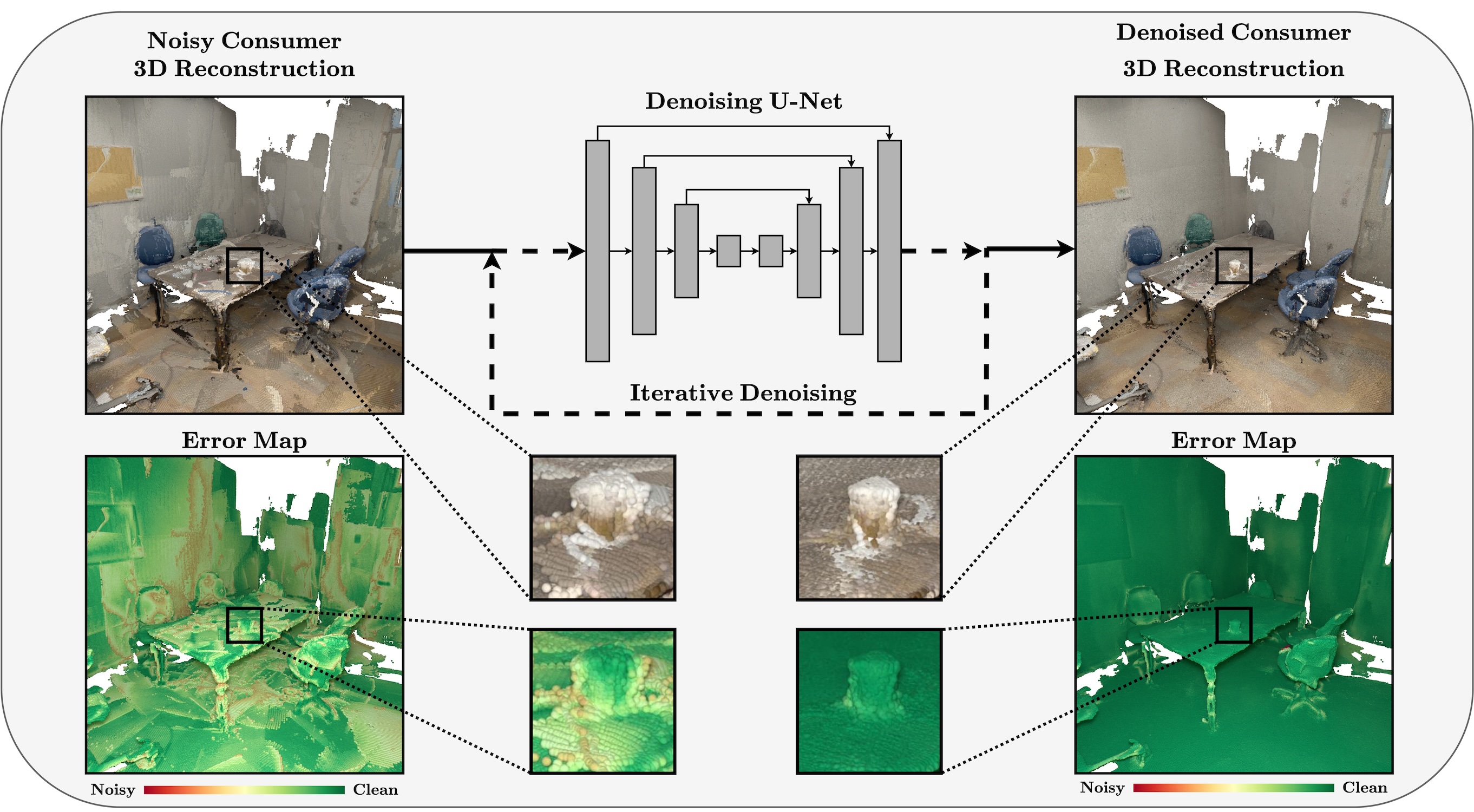}
\caption{Illustration of \name{} applied to a noisy LIDAR scan.}
\label{fig:method}
\end{figure}

\section{Introduction}
\label{sec:intro}

The use of point clouds to represent 3D objects and scenes \cite{Huang2023Segment3D, human3d,delitzas2024scenefun3d} is widespread across various fields, including 3D vision, robotics~\cite{zurbrugg2024icgnet, lemke2024spotcompose}, augmented/virtual reality, and autonomous driving~\cite{yilmaz24mask4d, kreuzberg2022stop}.
Recently, 3D scanning devices such as LIDAR sensors have gained popularity and have been incorporated into off-the-shelf consumer products.
These handheld devices allow users to scan objects and spaces in a relatively short amount of time.
However, the resulting point clouds often contain substantial noise due to hardware limitations, such as low scanner resolution, sensor noise, and limited range, or are affected by environmental factors like reflections, scattering, or occlusions.
Such noise can negatively impact downstream tasks that rely on high-quality point clouds 
\cite{yue2023connecting, Takmaz2023NIPS, engelmann2024opennerf}.
To address this issue, point cloud denoising has emerged as a critical technique for reducing noise in scanned objects and enhancing geometric details.

While significant advancements have been made in point cloud denoising research, cleaning up scans corrupted by real-world scanner noise remains challenging due to the need to accurately capture the underlying topology and nature of the clean surface, as well as the characteristics of the noise.
Although conventional point cloud denoising methods \cite{han2017review_filter, chen2018denoising_filter, zhang2020pointfilter, zheng2017guided_gf, digne_bilateral, zheng2018rolling_gf, han2018guided_gf, liu2020feature_gf, yadav2018constraint_gf} can perform well in specific circumstances, they often require extensive parameter fine-tuning or additional point features such as normals, and typically struggle to generalize to complex noise patterns.

Deep learning approaches \cite{rakotosaona2020pointcleannet, score_denoise, luo2020differentiable, zhao2023point, mao2022pd}  have demonstrated superior performance over traditional methods due to their data-driven nature.
One class of deep learning approaches \cite{luo2020differentiable, hermosilla2019total} tackles denoising by first resampling the point cloud to a coarse set of points, potentially eliminating high-frequency noise, and then recovering the underlying clean surface through upsampling and refinement.
Other methods \cite{duan20193d, zhang2020pointfilter,rakotosaona2020pointcleannet} aim to recover clean data via regression or point-wise displacement prediction, with PointCleanNet \cite{rakotosaona2020pointcleannet} also incorporating outlier removal.
Recently, score-based \cite{score_denoise, zhao2023point} and flow-based \cite{mao2022pd} models have shown promising results by learning the score or probability of the noise distribution directly. 
However, current models addressing noise in point clouds are typically trained under the assumption of synthetic noise, such as isotropic Gaussian noise.
Our experiments indicate that this assumption often falls short when denoising real-world 3D scans obtained from off-the-shelf devices, as it overlooks effects such as clusters of outliers, ghost points, or edge flares \cite{chen2018denoising_filter}.
Moreover, previous methods are often trained to minimize distance metrics that scale worse than linearly with the size of the input \cite{ravi2020accelerating}, hindering the scalability of model architectures—an essential factor in point-feature learning \cite{qian2022pointnext}.
Finally, recent models often focus on learning denoising tasks from point-based features like colors or normals only, which do not account for the high-level semantic properties of the underlying data \cite{oquab2023dinov2, Yue24p2Improving}.
To address these limitations, we propose incorporating high-level learned features by using point-wise features extracted from DINOv2 \cite{oquab2023dinov2}.

\clearpage

This work proposes a novel supervised approach for point cloud denoising based on diffusion models \cite{ddpm, DeBortoli2021DiffusionModeling}.
We approach the denoising task by formulating it as a Schrödinger bridge problem \cite{Schrödinger1932, DeBortoli2021DiffusionModeling}, solving it by training a network to find an optimum transport plan between the noisy and corresponding clean point cloud. This formulation allows to be trained on various types of data, including indoor scenes (\cf \cref{fig:method}).
We incorporate RGB and DINOV2 \cite{oquab2023dinov2} features to further enhance our method. Experiments show that our approach outperforms other state-of-the-art methods both in synthetic and real-world scenarios.

In summary, our main contributions are:

\begin{enumerate}
\item We propose \name{} (Pointcloud-to-Pointcloud Bridge), a novel approach for point cloud denoising inspired by the Schrödinger bridge problem, which formulates point cloud denoising as a tractable data-to-data diffusion process.
\item Furthermore, we advocate for semantically informed denoising by incorporating high-level features, such as DINOV2, to guide the denoising process.
\end{enumerate}

\section{Related Works}
\parag{Traditional denoising methods} can be roughly categorized into filter-based and optimization-based approaches. The filter-based methods draw from image and signal processing, assuming that the clean point cloud is corrupted with high-frequency noise.
Bilateral filter approaches \cite{han2017review_filter, chen2018denoising_filter, zhang2019point_filter} are suitable for denoising object surfaces while preserving sharp edges. Guided filter-based approaches \cite{han2018guided_gf, liu2020feature_gf, zheng2017guided_gf, zheng2018rolling_gf, yadav2018constraint_gf} attempt to fit a local linear model to a noisy point cloud aiming to preserve local details by using guidance from point coordinates or normals. Graph-based methods \cite{zeng20193d_gb, hu2020feature_gb, duan2018weighted_gb} model the point cloud as a graph to capture the underlying geometric structure and the relation of points with each other. Optimization-based methods range from sparse reconstruction \cite{leal2020sparse, digne2017sparse, mattei2017point_sparsity} to non-local-based point cloud denoising methods \cite{lu2020low_nonlocal, zeng20193d_gb, chen2019multi_nonlocal}. However, all these traditional methods rely on manually tuned hyper-parameters, which are tedious to obtain and typically do not generalize well.
\parag{Deep-learning-based methods} have recently shown promising results and improvement over traditional denoising methods.
PointCleanNet \cite{rakotosaona2020pointcleannet} first removes outliers and then predict point-wise displacement vectors to denoise the point clouds.
TotalDenoising \cite{hermosilla2019total} is an unsupervised method that applies total least squares \cite{leastsquares} to unstructured data such as point clouds. DMRDenoise \cite{luo2020differentiable} uses downsampling by differential pooling to estimate a manifold from which it resamples points to obtain a denoised point cloud.
Score-matching-based methods \cite{song2019generative} learn the score function of a tractable noise distribution and use (momentum) gradient ascent \cite{score_denoise, zhao2023point} to predict local displacements during inference. PD-Flow \cite{mao2022pd} utilizes normalizing flows to estimate the noise probability density function directly by disentangling the noise from the clean point cloud in a latent space. I-PFN \cite{de2023iterativepfn} improves upon iterative denoising methods using separated iteration modules for each denoising iteration already during training.

All mentioned methods except PD-Flow are all trained under the assumption of Gaussian noise since it is easy to generate training pairs of clean and noisy point clouds. However, as we will show experimentally, this does not necessarily translate to complex noise in real-world indoor scenes.
The main difference between our method and most previous works is that our method can be applied to any general data-to-data problem. By learning data-specific noise characteristics, our method better recovers the underlying clean data, removing outlier clusters and recovering fine details. Lastly, our method performs denoising using DDPM sampling \cite{ddpm}, making it more robust to the number of denoising steps as ablation studies show. Our method shows good results with as few as three function evaluations.
\parag{3D reconstruction} involves creating a three-dimensional representation of real-world scenes using 2D images and additional data such as depth. It differs from 3D point cloud denoising, but can serve as the initial step in generating 3D point clouds from real-world scenes. As a result, we will be discussing some relevant works in this field.
3DMatch \cite{zeng20163dmatch} is a data-driven approach for matching RGB-D reconstructions using learned volumetric features.
RoutedFusion~\cite{weder2020routedfusion} and Map-Adapt~\cite{zheng2024map} introduce machine learning-based approaches for real-time depth map fusion, and uses a neural network to predict non-linear updates for voxel-based fusion, addressing common errors and artifacts, especially for thin objects and edges.
NICE-SLAM \cite{Zhu2022CVPR} is a hierarchical grid-based SLAM method that uses RGB-D data for accurate environment reconstruction. It incorporates pre-trained models to enhance spatial understanding, improving mapping and tracking efficiency.
NICER-SLAM \cite{Zhu2023NICER} uses RGB input data to optimize an end-to-end joint mapping and tracking system, enabling it to predict colors, depths, and normals. Additional losses, such as warping and optical flow loss, further enhance the geometric consistency of NICER-SLAM.

\section{Method}
\label{sec:method}

\subsection{Overview}

Let the distributions of noisy point sets be denoted as  $\mathcal{\tilde{P}} = \{ \tilde{x}_{i}\} \in \mathbb{R}^{N \times D}$ and clean point sets as $\mathcal{P} = \{ x_{i}\} \in \mathbb{R}^{M \times D}$, where $D$ is the point feature dimension and $N$ and $M$ are the number of points in the noisy and clean point clouds, respectively.
We aim to denoise the noisy point sets $\mathcal{\tilde{P}}$ by leveraging diffusion models \cite{ddpm, Song2021DENOISINGMODELS, ddpm_pc}, exploiting their ability to predict clean data from noisy priors. 
Diffusion models have been successfully applied in various image generation and translation tasks
\cite{improved_ddpm, palette, i2sb} and more recently in point cloud generation and completion \cite{tyszkiewicz2023gecco, zeng2022lion, pvd, zeng2022lion, pc_completion_pretrained_tti, lyu2022a}. 
While most diffusion-based methods, as well as related works on point cloud denoising \cite{rakotosaona2020pointcleannet, score_denoise, zhao2023point} use Gaussian priors,
we argue that a data-to-data approach, rather than a data-to-noise approach,
is more suitable for point cloud denoising, especially when dealing with specific sensor data.
By using the distribution of noisy point sets $\mathcal{\tilde{P}}$ as the prior distribution, our method can learn data-dependent real-world noise characteristics.
However, training a diffusion model typically requires the process of diffusing a clean sample to a noisy sample to be tractable, as diffusion models are trained to learn step-wise noise removal. 
This process is generally unknown for real-world data.
A method to simulate a diffusion process between clean and noisy samples is the Schrödinger bridge (SB)
Schrödinger bridges have been increasingly utilized in generative models for tasks such as image-to-image translation \cite{DeBortoli2021DiffusionModeling, i2sb, Chen2022LIKELIHOODTHEORY}, protein matching \cite{somnath2023aligned}, and, more recently, text-to-speech \cite{Chen2023SchrodingerSynthesis}.
To our knowledge, we are the first to apply this approach to point cloud denoising.

\subsection{Pointcloud-to-Pointcloud Bridges}
\begin{figure}[tb]
    \centering
    \includegraphics[width=\textwidth]{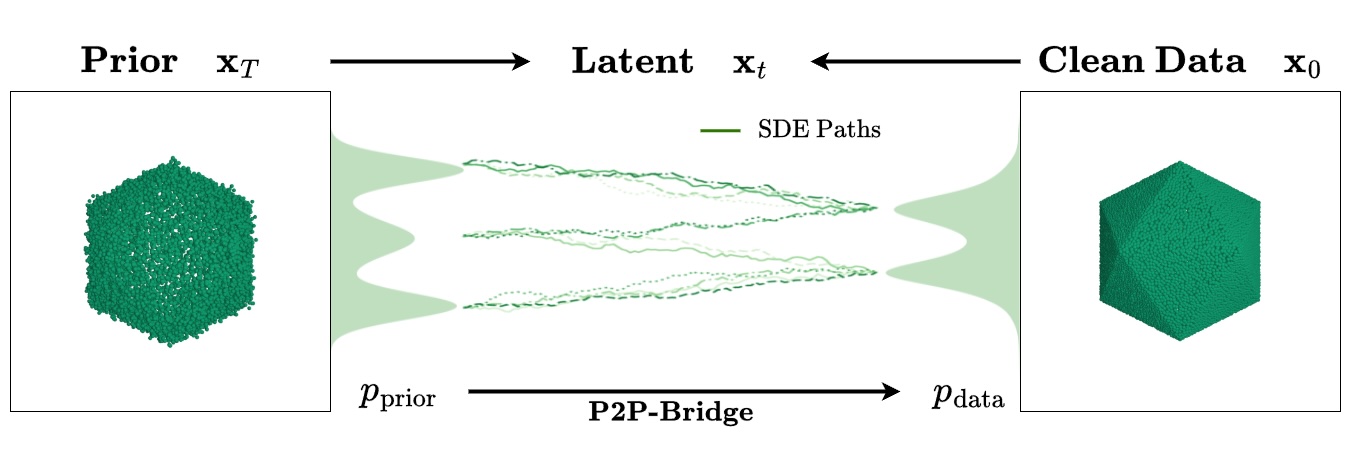}
    \caption{Illustration of \name{}, modeling point cloud denoising as a reverse data-to-data diffusion process. Our model can effectively transform noisy data into cleaner data by learning a bridge between clean and noisy data.}
    \label{fig:training}
\end{figure}

\parag{Tractable diffusion bridges.}
We approach the point cloud denoising problem from the perspective of diffusion models \cite{ddpm}.
By generating a diffusion process over $T$ timesteps $\{\mbf{x}{1}, \dots, \mbf{x}{T}\}$ with $\mbf{x}{t} \in \mathbb{R}^{N \times 3}$, a sample from clean data $\mbf{x}{0} \sim p_{\text{data}}$ is diffused into a noisy sample $\mbf{x}{T} \sim p_{\text{prior}}$.
In the context of point cloud denoising, the prior distribution corresponds to the distribution over noisy point sets $\mathcal{\tilde{P}}$ (\cf \cref{fig:training}).
Considering a reference path measure $p_{\text{ref}}(\mbf{x}{0:T})$ that describes this process, our objective is to find a process $p^*(\mbf{x}{0:T})$ such that $p^*(\mbf{x}{0}) =  p_{\text{data}}$ and $p^*(\mbf{x}{T}) =  p_{\text{prior}}$ minimizing the Kullback-Leibler divergence between $p_{\text{ref}}$ and $p^*$. This problem is also known as Schrödinger's bridge (SB) problem \cite{Schrödinger1932, leonard2013survey}.
It can be described using the forward and backward stochastic differential equations (SDEs) defined as follows:
\begin{align}
\begin{split}
    \label{eq:sde-schroedinger}
    \text{d} \mbf{x}{t} &= [\mathbf{f}(\mbf{x}{t}, t)\text{d}t + g^2(t) \nabla \log{\Psi_t(\mbf{x}{t})}]\text{d}t + g(t)\text{d}\mathbf{w}_t, \quad \mbf{x}{0} \sim p_{\text{data}}\\
    \text{d} \mbf{x}{t} &= [\mathbf{f}(\mbf{x}{t}, t)\text{d}t - g^2(t) \nabla \log{\hat{\Psi}_t(\mbf{x}{t})}]\text{d}t + g(t)\text{d}\mathbf{\bar{w}}_t, \quad \mbf{x}{t} \sim p_{\text{prior}}
\end{split}
\end{align}
where $\mathbf{w_t}$ is a Wiener process, $\mathbf{f}$ is a vector-valued function known as the \emph{drift} and $\mathbf{g}$ is a scalar-valued term referred to as the \emph{diffusion} coefficient. 
The terms $\nabla\log{\Psi_t(\mbf{x}{t})}$ and $\nabla\log{\hat\Psi_t(\mbf{x}{t})}$ are additional nonlinear drift terms that solve the following coupled partial differential equations (PDEs):
\begin{align}
\begin{split}
    \label{eq:couple_sb_sdes}
    \begin{cases}
    \frac{\delta \Psi}{\delta t} &= - \nabla_x \Psi^{\text{T}} \mathbf{f} - \frac{1}{2} \text{Tr}(g^2\nabla_x^2 \Psi) \\
    \frac{\delta \hat\Psi}{\delta t} &= - \nabla_x \hat\Psi^{\text{T}} \mathbf{f} + \frac{1}{2} \text{Tr}(g^2\nabla_x^2 \hat\Psi)
    \end{cases}
\end{split}
\end{align}
such that $\Psi_0\hat\Psi_0 = p_{\text{data}}, \Psi_T\hat\Psi_T = p_{\text{prior}}$ and $p_t=\Psi_t\hat\Psi_t$.
Chen \etal \cite{Chen2022LIKELIHOODTHEORY} show that \cref{eq:sde-schroedinger} is a generalization of score-based generative modeling (SGM) \cite{Song2021DENOISINGMODELS} to nonlinear processes.
Directly solving the system of differential equations is not practicable and is computationally expensive.
However, recent works \cite{i2sb, Chen2023SchrodingerSynthesis} introduce simplified, tractable frameworks under the assumption that we are dealing with paired boundary data, \ie, $p(\mbf{x}{0}, \mbf{x}{T}) = p_{\text{data}}(\mbf{x}{0})p_{\text{prior}}(\mbf{x}{T} \mid \mbf{x}{0})$.
In the context of point clouds, this means that the distribution over noisy point clouds is modeled as a joint distribution of clean point sets ($ p_{\text{data}}(\mbf{x}{0})$) and noise ($p_{\text{prior}}(\mbf{x}{T} \mid \mbf{x}{0})$).
When the boundary data is treated as a mixture of Diracs $(\delta_{\mbf{x}{0}}, \delta_{\mbf{x}{T}})$ with $\mathbf{f}:=0$ and using a linear diffusion schedule $g^2(t)$, it can be shown \cite{Chen2023SchrodingerSynthesis} that the posterior of \cref{eq:sde-schroedinger} has an analytic form given by:

\begin{equation}
    \label{eq:posterior_p2sb}
    q(\mbf{x}{t} \mid \mbf{x}{0}, x_T) = \mathcal{N}(\mbf{x}{t}; \mu_t(\mbf{x}{0}, \mbf{x}{T}), \Sigma_t)
\end{equation}
with
\begin{equation}
\label{eq:posterior_interpolation}
    \mbf{\mu}{t} = \frac{\sigbts}{\sigbts + \sigts} \mbf{x}{0} + \frac{\sigts}{\sigbts + \sigts} x_T \quad \text{and} \quad \Sigma_t = \frac{\sigts \sigbts}{\sigts + \sigbts}
\end{equation}
where $\sigts = \int_0^t g^2(\tau)\text{d}\tau$ and $\sigbts = \int_t^1 g^2(\tau)\text{d}\tau$. This simplifies \cref{eq:couple_sb_sdes} and makes it fully tractable. 
We can parameterize a network $\epsilon_{\theta}$ that predicts the noise added to $\mbf{x}{0}$ at timestep $t$ resulting in the noisy sample $\mbf{x}{t}$ using the noise-prediction loss:
\begin{equation}
    \label{eq:p2sb_loss}
    \mathcal{L} = \|\epsilon_\theta(\mbf{x}{t}, t) - \frac{\mbf{x}{t} - \mbf{x}{0}}{\sigma_t}\|_2^2.
\end{equation}
During inference, we can iteratively sample using DDPM sampling \cite{ddpm}
\begin{equation}
    \label{eq:bridge_sampling}
    p(\mbf{x}{t-1} \mid \mbf{x}{t},\mbf{\hat{x}}{0}) = \mathcal{N}(\mbf{x}{t}; \mbf{\mu}{t}(\mbf{\hat{x}}{0}, \mbf{x}{T}), \Sigma_t), \quad \mbf{\hat{x}}{0} = \mbf{x}{t} - \sigma_t \epsilon_\theta(\mbf{x}{t}, t),
\end{equation}
as this induces the same marginal density of SB paths as long as $\mbf{\hat x}{0}$ is close to the actual $\mbf{x}{0}$ \cite{i2sb, Chen2023SchrodingerSynthesis}.
However, sampling from $\mbf{\mu}{t}$ describes an interpolation between two point clouds.
This operation is straightforward for data with a fixed domain, such as image data, where pixels are attached to a static grid.
It is, however, not well defined for unordered point clouds and depends on a proper metric \cite{pointmixup}.

\parag{Meaningful interpolation between unordered point sets.}
We use the shortest path interpolation method from PointMixup \cite{pointmixup} to describe the path of the posterior mean $\mbf{\mu}{t}$. Shortest path interpolation tries to find an optimum assignment $\phi^*$ that minimizes the average distance for each point in the point cloud $\mbf{x}{T}$ to its nearest neighbor in $\mbf{x}{0}$. Assuming that the noisy and clean point sets both contain $N$ points, the assignment problem is defined as:
\begin{equation}
    \phi^* = \operatorname*{arg\,min}_{\phi \in \mbf{\Phi}} \sum_{i=1}^N \| \mbf{x}{T}^{i} - \mbf{x}{0}^{\phi(i)} \|_2,
\end{equation}
where $\mbf{\Phi} = \{\{1, \dots, N\} \rightarrow \{1, \dots, N\}\}$ is the set of possible bijective assignments between points in $\mbf{x}{T}$ and $\mbf{x}{0}$.
Using shortest-path-interpolation resembles finding an optimal transport plan between two point sets when the cost is a squared geodesic distance \cite{Peyre2019ComputationalTransport}.
The resulting path from shortest-path-interpolation corresponds to the path taken by the posterior of \cref{eq:sde-schroedinger} when the stochasticity of the bridge vanishes, \ie{}, $g^2(t) \rightarrow 0$ \cite{Peyre2019ComputationalTransport, i2sb, Chen2023SchrodingerSynthesis} which motivates the choice of nearest-path-interpolation over other possible interpolation methods. Diminishing the stochasticity of the bridge effectively reduces the bridge SDE to an optimal transport ordinary differential equation (OT-ODE) of the form:
\begin{equation}
    \text{d}\mbf{x}{t} = \frac{g^2(t)}{\sigts} (\mbf{x}{t} - \mbf{x}{0}) \text{d}t.
\end{equation}
In practice, we have to calculate the optimal assignment for every data pair in our dataset only once. Subsequently, we can employ $\phi^*$ to reorder the clean point clouds so that they are aligned with their corresponding noisy point clouds. During training, we can sample $\mbf{x}{t}$ without solving the optimum assignment problem again, allowing fast and scalable training. Further discussion and experiments on shortest-path interpolation can be found in the appendix.

\subsection{Implementation}

\begin{figure}[b!]
\centering
\includegraphics[width=\linewidth]{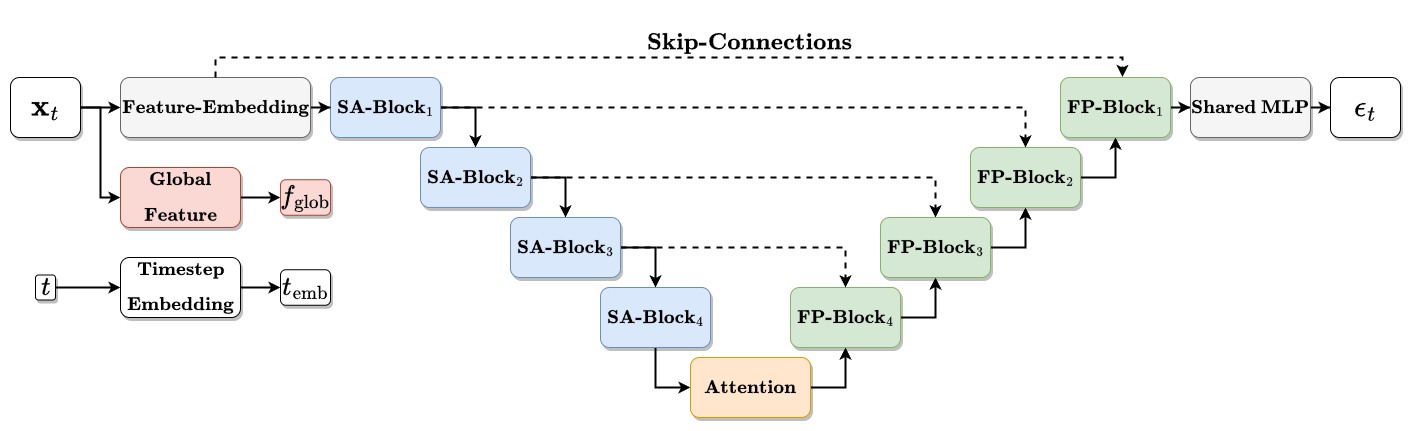}
\caption{Our network architecture is based on PointVoxelConvolutions (PVC) \cite{pvcnn}.
We adapt the network implementation from LION \cite{zeng2022lion},
augmenting it with multi-headed global attention and a feature embedding module.
Both the feature embedding and the final shared MLP block are implemented using $1 \times 1$ convolutions.
}
\label{fig:architecture}
\end{figure}

\parag{Model architecture.}
We follow previous works on point cloud diffusion models \cite{pvd, zeng2022lion, lyu2022a} and utilize a model architecture (\cf \cref{fig:architecture}) based on PointVoxel-CNN (PVCNN) \cite{pvcnn}.
PVCNN is a PointNet++ \cite{qi2017pointnet++} inspired architecture that enhances the set abstraction (SA) and feature propagation (FP) blocks with global features extracted from a voxelized point cloud representation.
Similar to LION \cite{zeng2022lion}, we incorporate squeeze-excitation blocks \cite{hu2018squeeze} and a global feature extraction network.
Additionally, we employ a feature embedding layer that maps the incoming features to a higher dimension using a $1 \times 1$ convolution.
We follow recent advances in image diffusion models \cite{hoogeboom2023simple} and apply attention only at the lowest layer, utilizing multiple attention heads.
The network is conditioned on timestep $t$ using sinusoidal positional embeddings and point features.
Global features are incorporated via adaptive group normalization.
The input data is processed in patches, sampling points from dataset-dependent radius spheres to ensure spatial correspondence between noisy and clean data.

\parag{Input features.}
Additional RGB data is often available from mobile phone scans.
We propose extracting point-wise features from the raw RGB data using DINOV2 \cite{oquab2023dinov2}.
The pixel-wise DINOV2 features are projected onto the noisy point cloud using camera poses and intrinsics, resulting in point-wise features.
\section{Experiments}
\label{sec:experiments}

\subsection{Datasets}

For evaluation, we compare our \name{} on well-established 3D object datasets depicting single objects and large-scale indoor scene datasets.
For object-level denoising, we use the PU-Net dataset \cite{yu2018pu}, which consists of 40 objects for training and 20 objects for evaluation.
Additionally, we use the PC-Net dataset \cite{rakotosaona2020pointcleannet} to provide another ten objects for evaluation only.
For both object datasets, we follow the standard practice of simulating noise using isotropic Gaussian distributions.

Unlike prior work, we also propose evaluating on scene-level point clouds, a setup that more closely reflects real-world usage scenarios.
For evaluation on real-world datasets, we select the indoor scene datasets ScanNet++ \cite{yeshwanthliu2023scannetpp} and ARKitScenes \cite{dehghan2021arkitscenes}, as they provide paired clean and noisy point cloud data along with pose information.
ScanNet++ contains 330 indoor scenes, where each scene includes a series of noisy depth maps captured by a handheld LiDAR scanner and a clean scan obtained by a Faro laser scanner.
We use the 3D reconstruction script provided by the authors of ScanNet++ to construct the noisy point clouds, on which we then apply the denoising methods.

Additionally, we evaluate performance when the reconstruction is refined using 3DMatch \cite{zeng20163dmatch}. 
ARKitScenes contains 5047 scans of various indoor venues, with noisy scans generated using Apple ARKit surface reconstruction and clean scans acquired by a Faro laser scanner.
Further details can be found in the appendix.

\subsection{Evaluation Metrics}
We use the Chamfer distance (CD) and the Point-to-Mesh (P2M) distance as quantitative evaluation metrics.
The CD measures the similarity between the predicted point cloud $\mathcal{\hat{P}} = \{\hat{x}_i \in \mathbb{R}^3\}_{i=1}^{n}$ and the corresponding clean point cloud $\mathcal{P} = \{x_j \in \mathbb{R}^3\}_{j=1}^{m}$, and is defined as:
\begin{equation}
\label{CD}
\text{CD}(\mathcal{\hat{P}}, \mathcal{P}) = \underbrace{\frac{1}{2n} \sum_{i=1}^{n} \| \hat{x}_i - \text{NN}(\hat{x}_i, \mathcal{P}) \|_2^2}_{\cdforw} + \underbrace{\frac{1}{2m} \sum_{j=1}^{m} \| x_j - \text{NN}(x_j, \mathcal{\hat{P}}) \|_2^2}_{\cdbakw},
\end{equation}
where NN is the nearest-neighbor function.
The first term, $\cdforw$, approximately describes the average distance from noisy points to the ground truth surface, while the second term, $\cdbakw$, encourages uniform coverage.
The Point-to-Mesh (P2M) distance is defined as follows:
\begin{equation}
\label{P2M}
    \text{P2M}(\mathcal{\hat{P}}, \mathcal{M}) = \underbrace{\frac{1}{2n} \sum_{i=1}^{n} \min_{f \in \mathcal{M}} d(\hat{x}_i,f)}_{\text{F2P}} + \underbrace{\frac{1}{2|\mathcal{M}|} \sum_{f \in \mathcal{M}}  \min_{\hat{x}_i \in \mathcal{\hat{P}}} d(\hat{x}_i,f)}_{\text{P2F}}
\end{equation}
where $d(x, f)$ is a function that measures the distance from point $x$ to face $f$.
The first term, therefore, represents the Face-to-Point distance (F2P), while the second term corresponds to the Point-to-Face distance (P2F).
For calculating object-level metrics, we center and scale the prediction and ground truth to the unit sphere following the procedure in ScoreDenoise \cite{score_denoise}.

\subsection{Experimental Details}
We train our model on the PU-Net dataset to denoise artificially noised objects with Gaussian noise, following previous works.
For ScanNet++ and ARKitScenes, we train all deep-learning-based denoising methods, including ours, with a batch size of 32 for up to 100,000 steps.
We use the training parameters and model weights provided in the publicly available code bases for previous works \cite{score_denoise, luo2020differentiable, zhao2023point, mao2022pd}.
Additional experimental details are provided in the appendix.

\subsection{Comparison on Objects}
We quantitatively evaluate our method against traditional approaches such as Bilateral \cite{digne_bilateral} or GLR \cite{zeng20193d_gb}, as well as deep-learning-based methods including PC-Net \cite{rakotosaona2020pointcleannet}, DMR \cite{luo2020differentiable}, ScoreDenoise \cite{score_denoise}, MAG \cite{zhao2023point} and PD-Flow \cite{mao2022pd}.
For the evaluation, we use Gaussian noise levels ranging from 1\% to 3\% of the object's bounding box diagonal for both sparse (10k points) and dense (10k points) objects.
\Cref{tab:object_scores} shows that our method outperforms previous optimization-based and deep-learning-based methods in most noise settings.
In the 1\% noise setting, our method performs second best to PD-Flow, while for higher noise levels, we observe a significant increase in accuracy compared to previous methods.
Our method also seems to adapt better to unseen objects, as indicated by the results on the PC-Net dataset.
Notably, these results are achieved with only three denoising steps.
\Cref{fig:pu_qualitative} provides a qualitative comparison of our method with recent deep-learning-based methods on objects denoised from 3\% isotropic Gaussian noise.
Our method seems to produce less noisy and smoother results than previous approaches.
Additional qualitative results, experiments with different noise types, and run-time analyses are provided in the supplementary materials.

\begin{table}[ht!]
\centering
\caption{
\textbf{Object-level Denoising Results.}
We show the Chamfer Distance (CD) and Point-2-Mesh (P2M) distance metrics on the PU-Net \emph{(top)} and PC-Net \emph{(bottom)} datasets.
All scores are multiplied by $10^4$ for readability.
When available, baseline scores are taken from ScoreDenoise~\cite{score_denoise} and MAG~\cite{zhao2023point}; otherwise, we use publicly available weights and testing scripts to evaluate on the test data provided by \cite{score_denoise}.
}

\resizebox{\columnwidth}{!}{%
\setlength{\tabcolsep}{3pt}
		
		\begin{tabular}{llcccccccccccc}%
			\toprule
			\multicolumn{2}{l}{Num. of Points} & \multicolumn{6}{c}{$10 \cdot 10^3$ (sparse)} & \multicolumn{6}{c}{$50 \cdot 10^3$ (dense)} \\
			\cmidrule(lr){3-8} \cmidrule(lr){9-14}
			\multicolumn{2}{l}{Gaussian Noise} & \multicolumn{2}{c}{1\%} & \multicolumn{2}{c}{2\%} & \multicolumn{2}{c}{3\%} & \multicolumn{2}{c}{1\%} & \multicolumn{2}{c}{2\%} & \multicolumn{2}{c}{3\%} \\
			\cmidrule(lr){3-4} \cmidrule(lr){5-6} \cmidrule(lr){7-8}
			\cmidrule(lr){9-10} \cmidrule(lr){11-12} \cmidrule(lr){13-14}
			                                                        & \textbf{Method}                             & {CD}              & {P2M}             & {CD}              & {P2M}             & {CD}               & {P2M}             & {CD}              & {P2M}             & {CD}              & {P2M}                                     & {CD}              & {P2M}              \\
			\midrule
			\multirow{7}{*}{\rotatebox{90}{PU-Net \cite{yu2018pu}}} & {Bilateral \cite{digne_bilateral}}          & 3.65              & 1.34              & 5.01              & 2.02              & 7.00               & 3.56              & 0.88              & 0.23              & 2.38              & 1.39                                      & 6.30              & 4.73               \\
			                                                        & {PCNet~\cite{rakotosaona2020pointcleannet}} & 3.52              & 1.15              & 7.47              & 3.97              & 13.1               & 8.74              & 1.05              & 0.35              & 1.45              & 0.61                                      & \rd 2.29          & 1.29               \\
			                                                        & {DMRDenoise~\cite{luo2020differentiable}}   & 4.48              & 1.72              & 4.98              & 2.12              & 5.89               & 2.85              & 1.16              & 0.47              & 1.57              & 0.80                                      & 2.43              & 1.53               \\
			                                                        & {GLR~\cite{zeng20193d_gb}}                  & 2.96              & 1.05              & 3.77              & 1.31              & 4.91               & 2.11              & 0.70              & 0.16              & 1.59              & 0.83                                      & 3.84              & 2.71               \\
			                                                        & {ScoreDenoise~\cite{score_denoise}}         & 2.52              & 0.46              & 3.69              & 1.07              & \rd 4.71           & \rd 1.94          & 0.72              & \rd 0.15          & \rd 1.29          & 0.57                                      & \nd 1.93          & \nd 1.04           \\
			                                                        & {MAG~\cite{zhao2023point}}                  & 2.50              & 0.46              & 3.63              & 1.05              & \nd 4.69           & \rd 1.92          & 0.71              & \rd 0.15          & \rd 1.29          & \rd 0.56                                  & \nd 1.93          & \rd 1.05           \\
			                                                        & {PD-Flow~\cite{mao2022pd}}                  & \st 2.13          & \nd 0.38          & \nd 3.25          & \nd 1.01          & 5.19               & 2.52              & \nd 0.65          & 0.16              & 1.42              & 0.78                                      & 3.90              & 2.86               \\
			                                                        & {I-PFN~\cite{de2023iterativepfn}}           & \rd2.31           & \st0.37           & \rd3.43           & \nd0.9            & 5.49               & 2.5               & \rd0.66           & \nd0.12           & \nd1.05           & \nd0.43                                   & 2.54              & 1.65               \\
			                                                        & \textbf{\name{} (Ours)}                     & \nd \tbf{2.28}    & \rd \tbf{0.39}    & \st \tbf{3.20}    & \st \tbf{0.81}    & \st \tbf{3.99}     & \st \tbf{1.42}    & \st \tbf{0.59}    & \st \tbf{0.09}    & \st \tbf{0.90}    & \st \tbf{0.32}                            & \st \tbf{1.56}    & \st \tbf{0.84}     \\
			\midrule
			\multirow{7}{*}{\rotatebox{90}{PC-Net \cite{rakotosaona2020pointcleannet}}}
			                                                        & {Bilateral~\cite{digne_bilateral}}          & 4.32              & 1.35              & 6.17              & 1.65              & 8.30               & 2.39              & 1.17              & 0.20              & 2.50              & 0.63                                      & 6.08              & 2.19               \\
			                                                        & {PCNet~\cite{rakotosaona2020pointcleannet}} & 3.85              & 1.22              & 8.75              & 3.04              & 14.5               & 5.87              & 1.29              & 0.29              & 1.91              & 0.51                                      & 3.25              & 1.08               \\
			                                                        & {DMRDenoise~\cite{luo2020differentiable}}   & 6.60              & 2.15              & 7.15              & 2.24              & 8.09               & 2.49              & 1.57              & 0.35              & 2.01              & 0.49                                      & \rd 2.99          & \rd 0.86           \\
			                                                        & {GLR~\cite{zeng20193d_gb}}                  & 3.40              &  0.96          & 5.27              & \rd 1.15          & 7.25               & \rd 1.67          & \nd 0.96          & \nd 0.13          & 2.02              & 0.42                                      & 4.50              & 1.31               \\
			                                                        & {ScoreDenoise~\cite{score_denoise}}         &  3.37          &  0.83          &  5.13          & 1.20              & \rd 6.78           & 1.94              & 1.07              & 0.18              & \nd 1.66          & \nd 0.35                                  & \nd 2.49          & \nd 0.66           \\
			                                                        & {MAG~\cite{zhao2023point}}                  &  3.37          &  0.83          &  5.13          & 1.19              & 7.24               & 1.94              & 1.07              & 0.18              & \nd 1.66          & \nd 0.35                                  & 3.56              & 1.15               \\
			                                                        & {PD-Flow~\cite{mao2022pd}}                  & \rd 3.24          & \st 0.62          & \nd 4.62          & \nd 0.92          & \nd 6.61           & \nd 1.62          & \rd 0.97          &  0.15          &  1.80          &  0.40                                  & 4.28              & 1.37               \\
			                                                        & {I-PFN~\cite{de2023iterativepfn}}           & \nd3.05              & \rd 0.72              & \rd 4.95              & 1.16              & 7.39               & 2.21              & 0.99              & \rd 0.14              & \nd 1.43              & \nd 0.27                                      & 3.03              & \rd 0.86               \\
			                                                        & \textbf{\name{} (Ours)}                     & \st \textbf{2.88} & \nd \textbf{0.63} & \st \textbf{4.47} & \st \textbf{0.89} & \st  \textbf{5.58} & \st \textbf{1.29} & \st \textbf{0.92} & \st \textbf{0.12} & \st \textbf{1.35} & \st \textbf{0.24}                         & \st \textbf{2.12} & \st  \textbf{0.49} \\
			\bottomrule
		\end{tabular}}
	\label{tab:object_scores}
\end{table}

\begin{figure}[ht!]

\begin{center}

\resizebox{0.5\textwidth}{!}{
    \begin{tabular}{rcl}
Noisy &\includegraphics[width=0.6\linewidth]{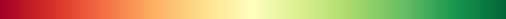} & Clean
    \end{tabular}
}
\\
\includegraphics[width=0.16\linewidth,trim={210 130 190 70},clip]{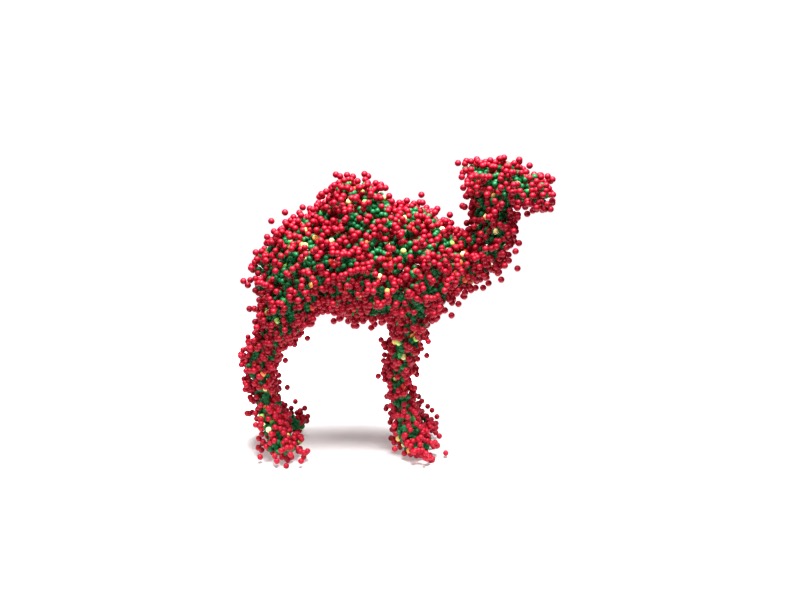}
\includegraphics[width=0.16\linewidth,trim={210 130 190 70},clip]{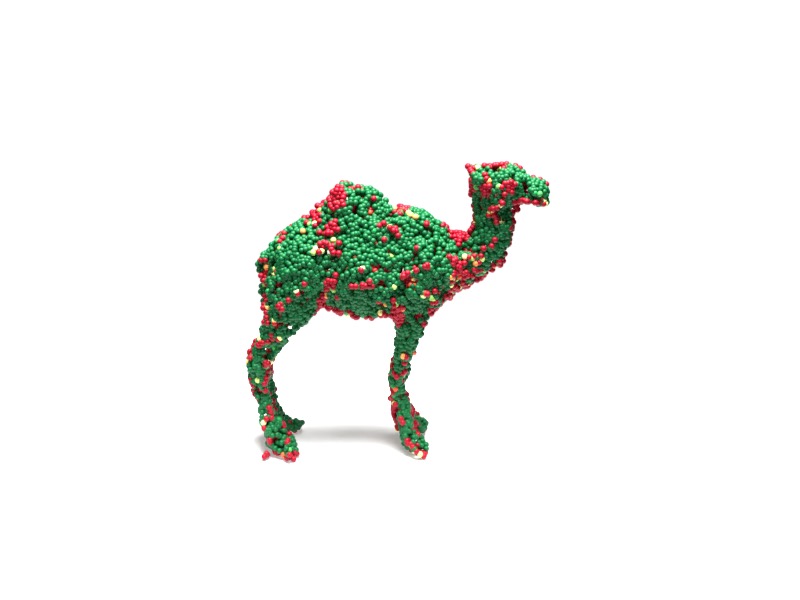}
\includegraphics[width=0.16\linewidth,trim={210 130 190 70},clip]{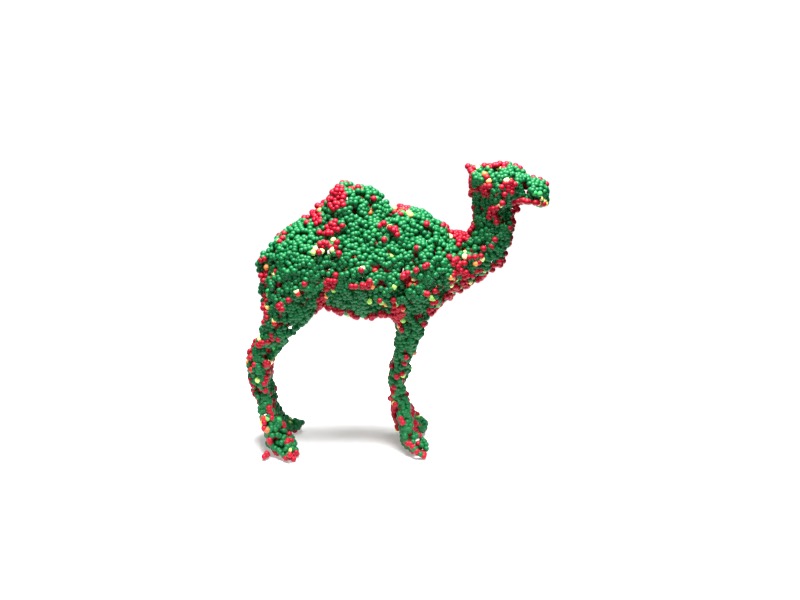}
\includegraphics[width=0.16\linewidth,trim={210 130 190 70},clip]{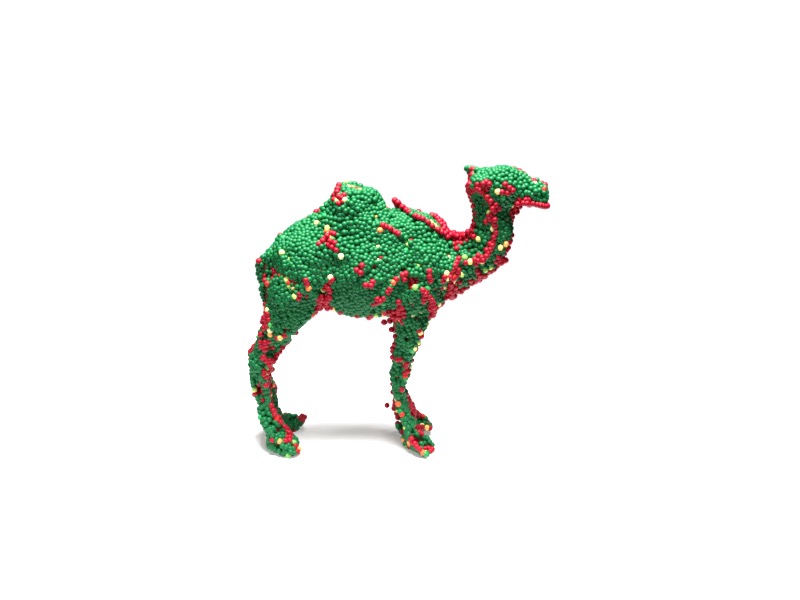}
\includegraphics[width=0.16\linewidth,trim={210 130 190 70},clip]{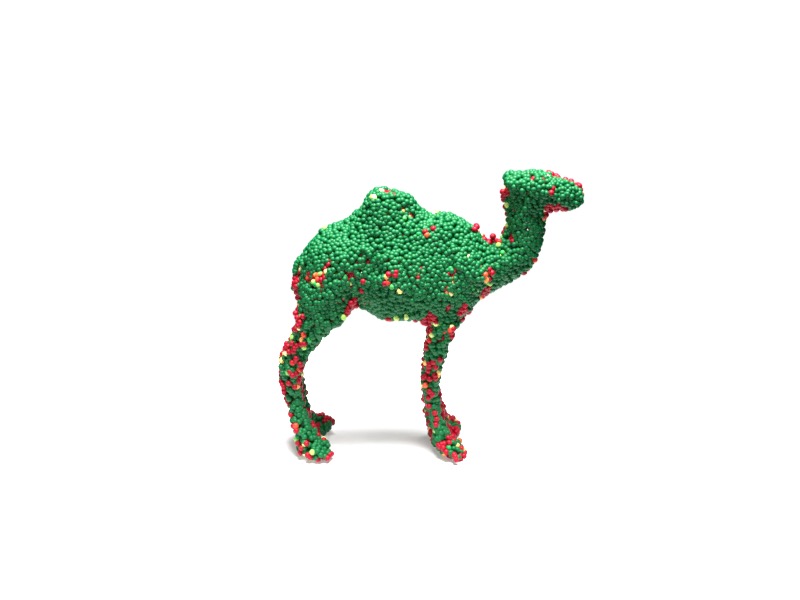}
\includegraphics[width=0.16\linewidth,trim={210 130 190 70},clip]{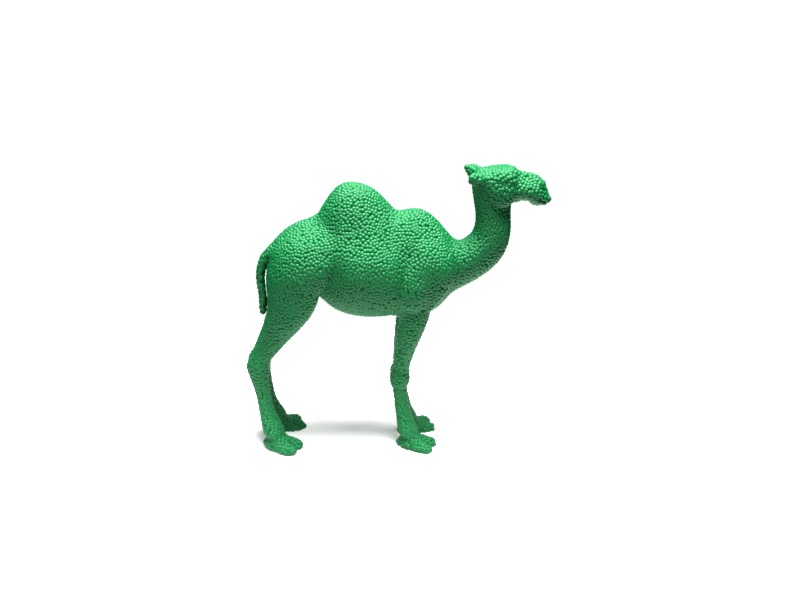}
\\
\includegraphics[width=0.16\linewidth,trim={210 110 190 70},clip]{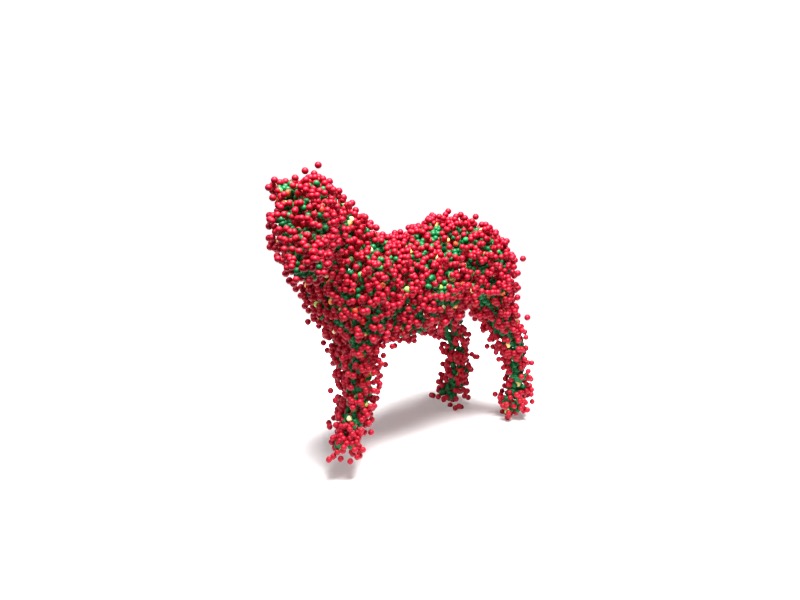}
\includegraphics[width=0.16\linewidth,trim={210 110 190 70},clip]{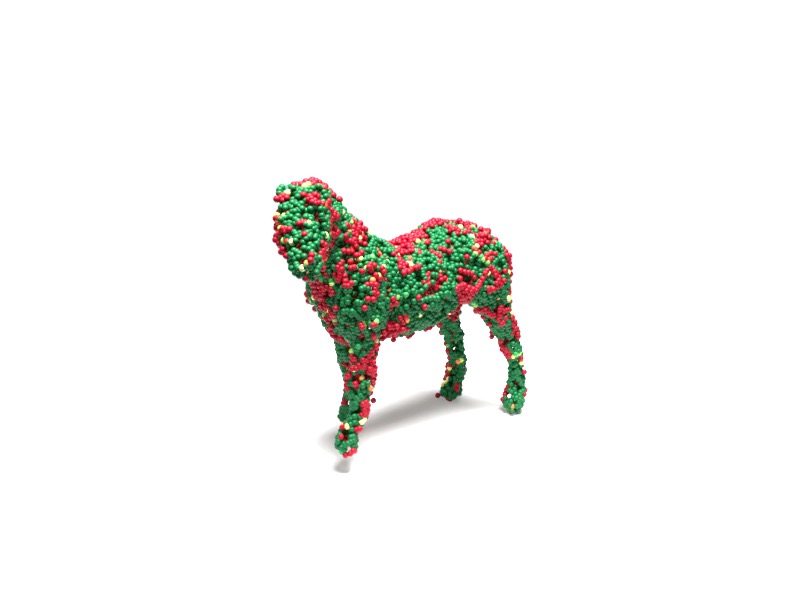}
\includegraphics[width=0.16\linewidth,trim={210 110 190 70},clip]{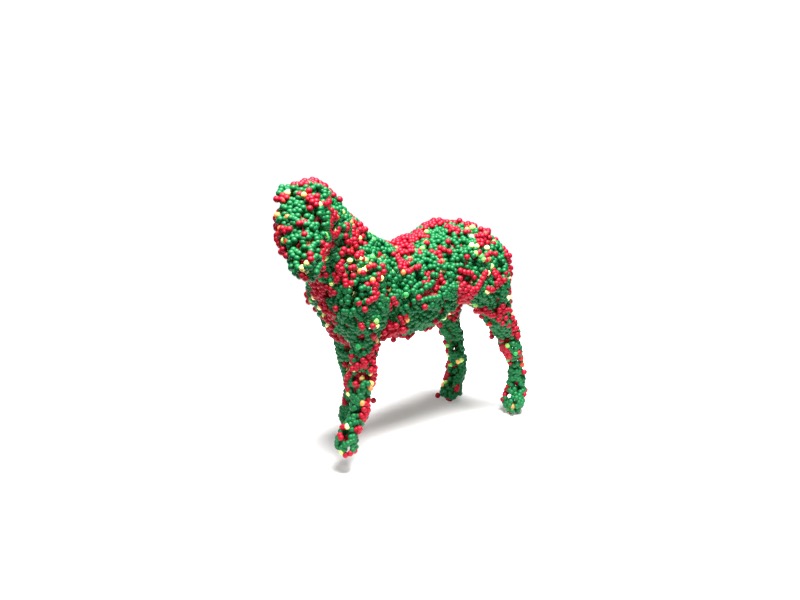}
\includegraphics[width=0.16\linewidth,trim={210 110 190 70},clip]{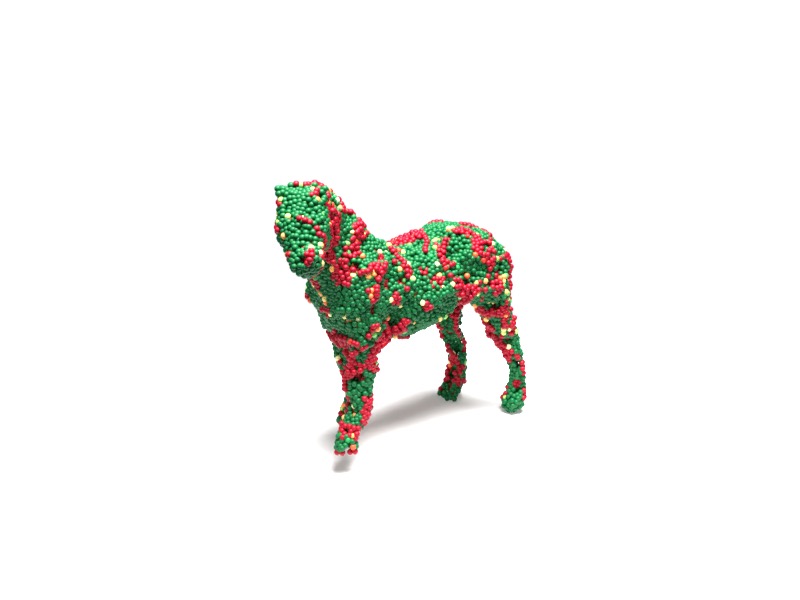}
\includegraphics[width=0.16\linewidth,trim={210 110 190 70},clip]{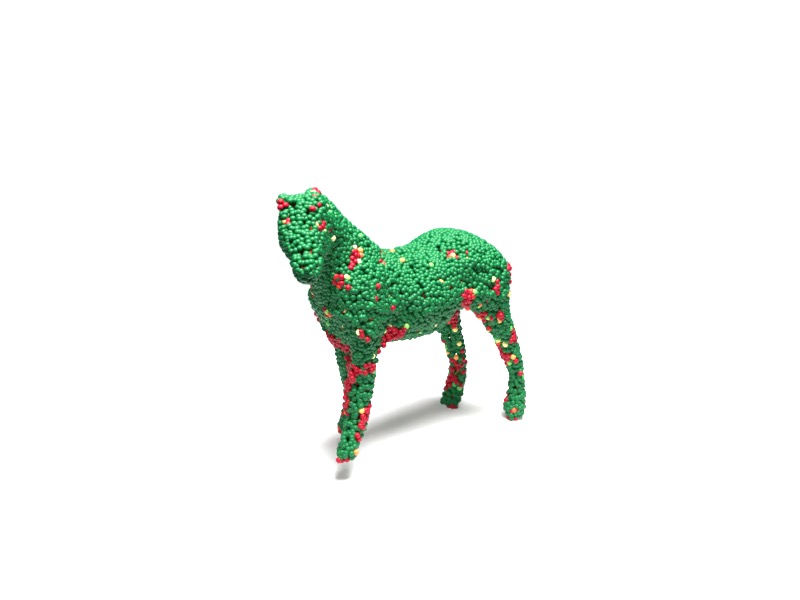}
\includegraphics[width=0.16\linewidth,trim={210 110 190 70},clip]{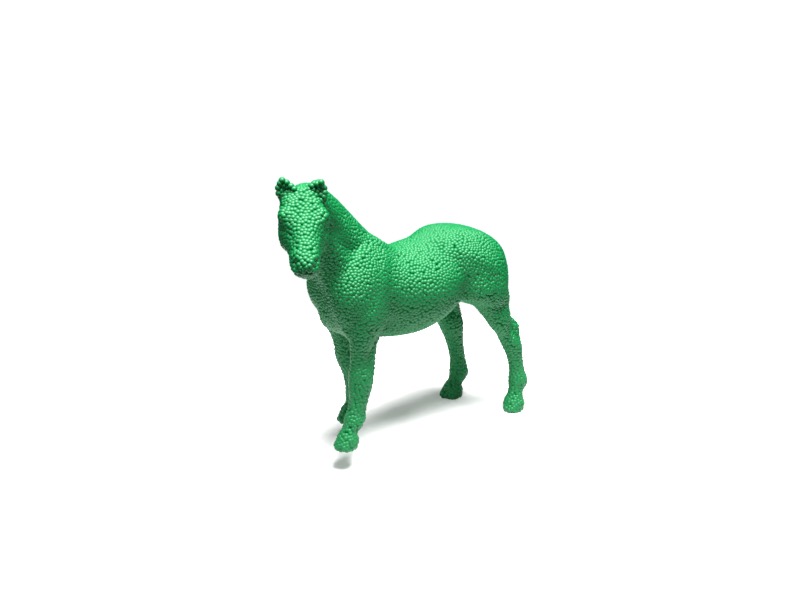}
\\
\includegraphics[width=0.16\linewidth,trim={190 120 150 20},clip]{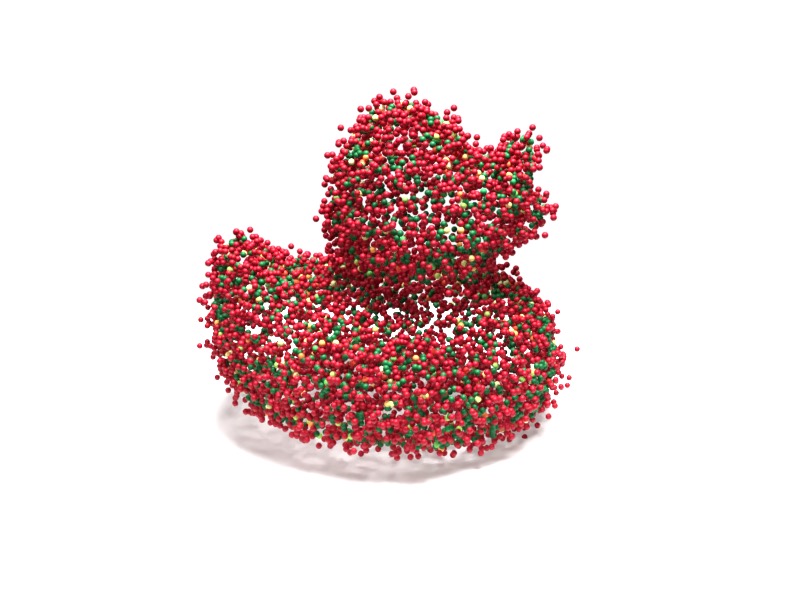}
\includegraphics[width=0.16\linewidth,trim={190 120 150 20},clip]{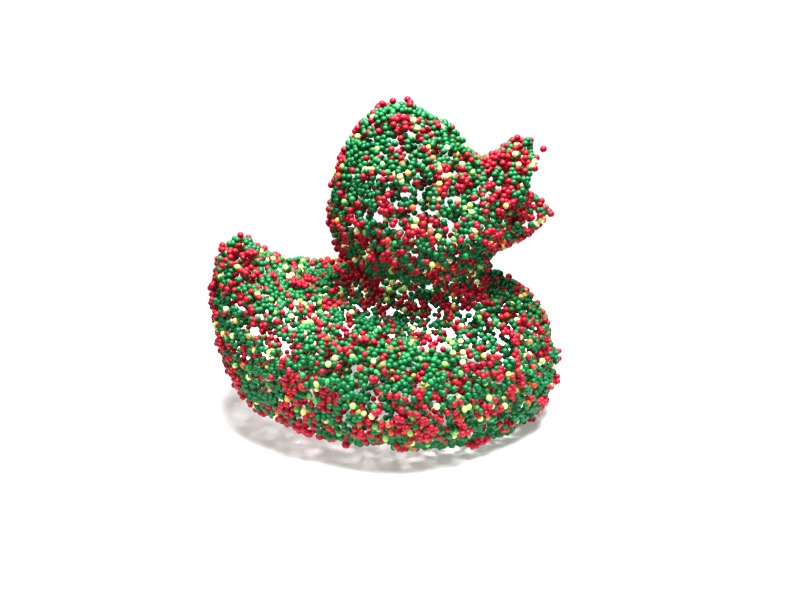}
\includegraphics[width=0.16\linewidth,trim={190 120 150 20},clip]{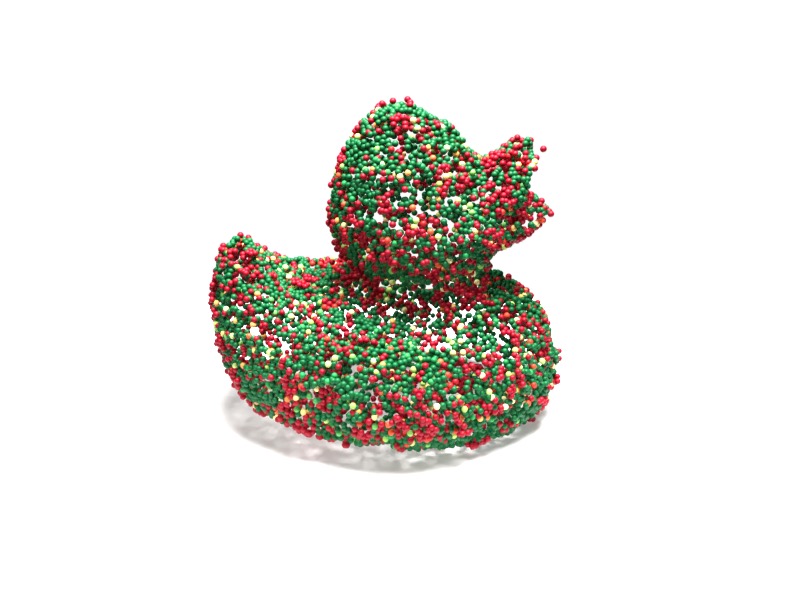}
\includegraphics[width=0.16\linewidth,trim={190 120 150 20},clip]{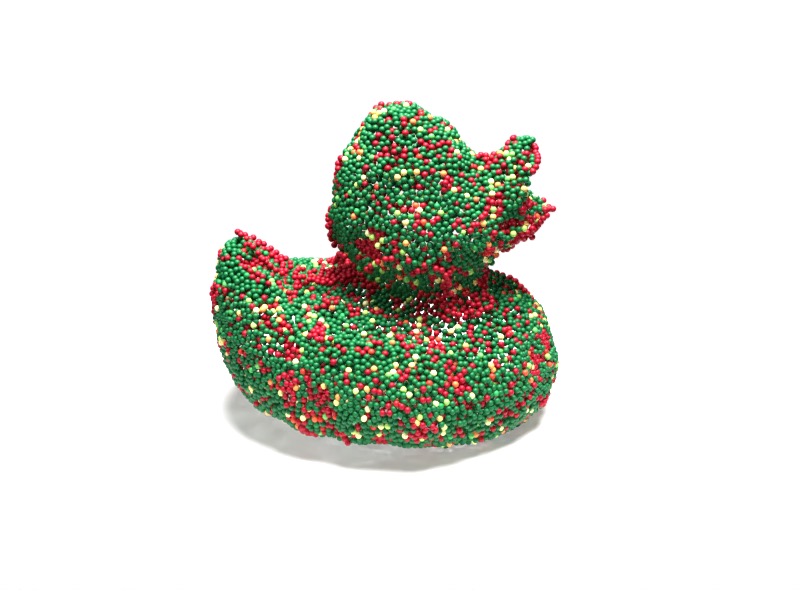}
\includegraphics[width=0.16\linewidth,trim={190 120 150 20},clip]{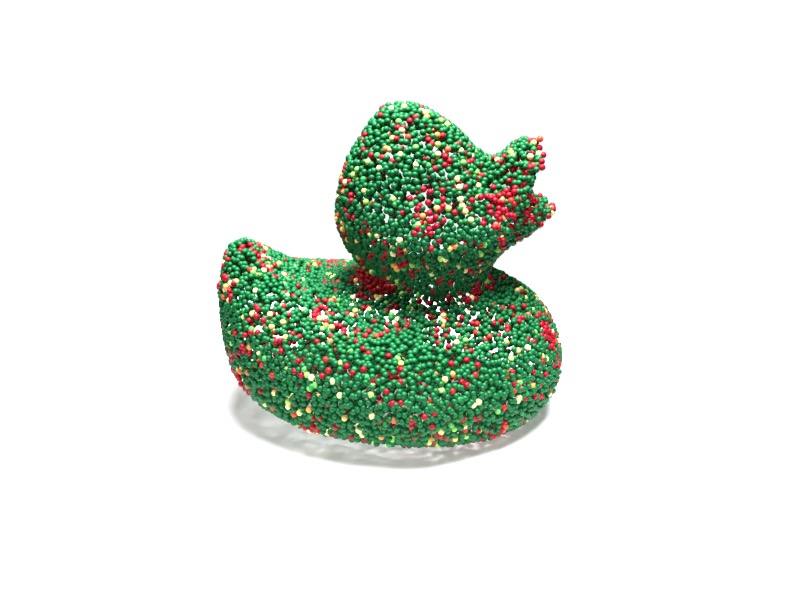}
\includegraphics[width=0.16\linewidth,trim={190 120 150 20},clip]{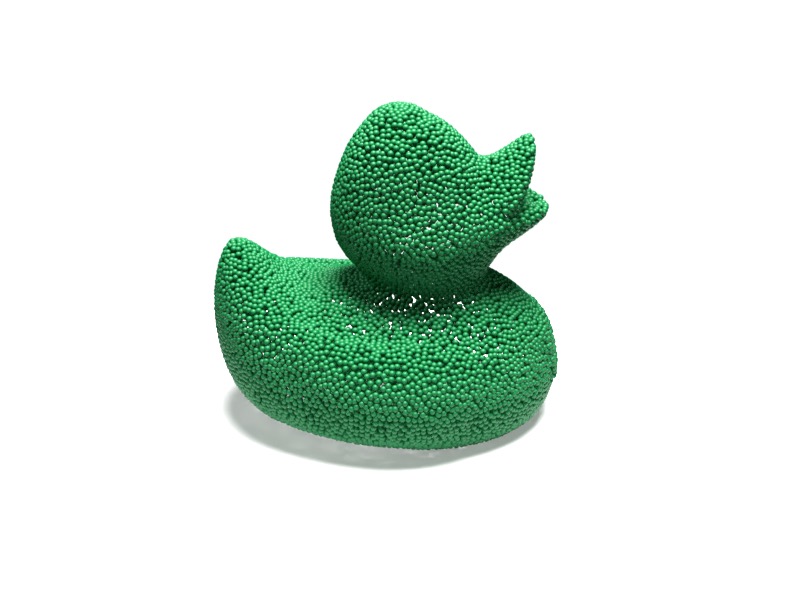}
\\
\includegraphics[width=0.16\linewidth,trim={210 110 190 30},clip]{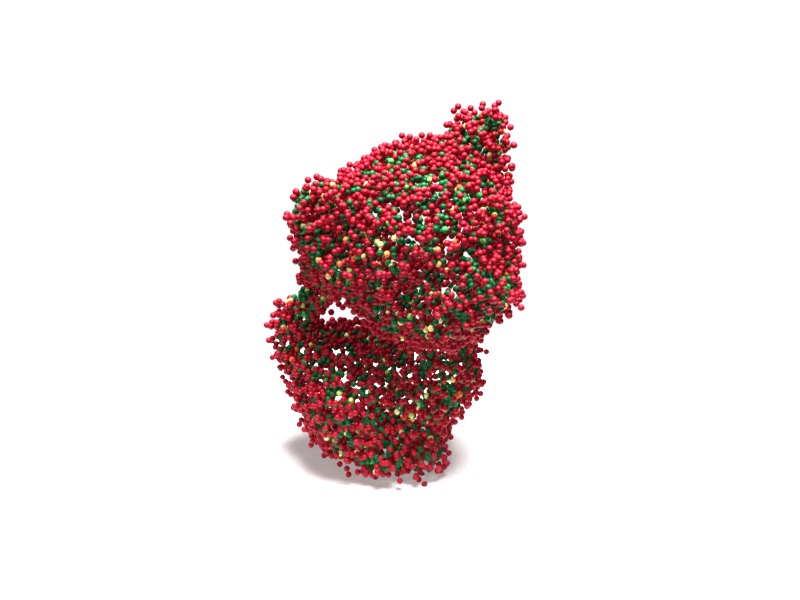}
\includegraphics[width=0.16\linewidth,trim={210 110 190 30},clip]{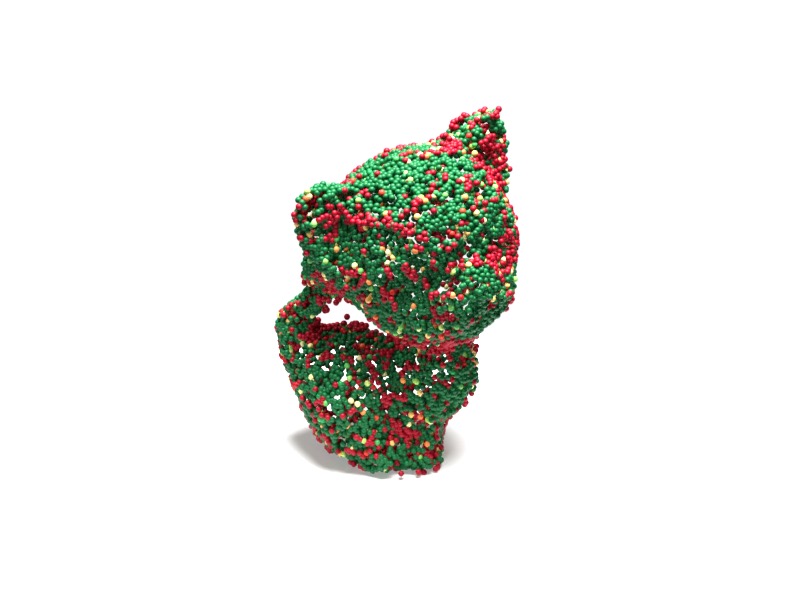}
\includegraphics[width=0.16\linewidth,trim={210 110 190 30},clip]{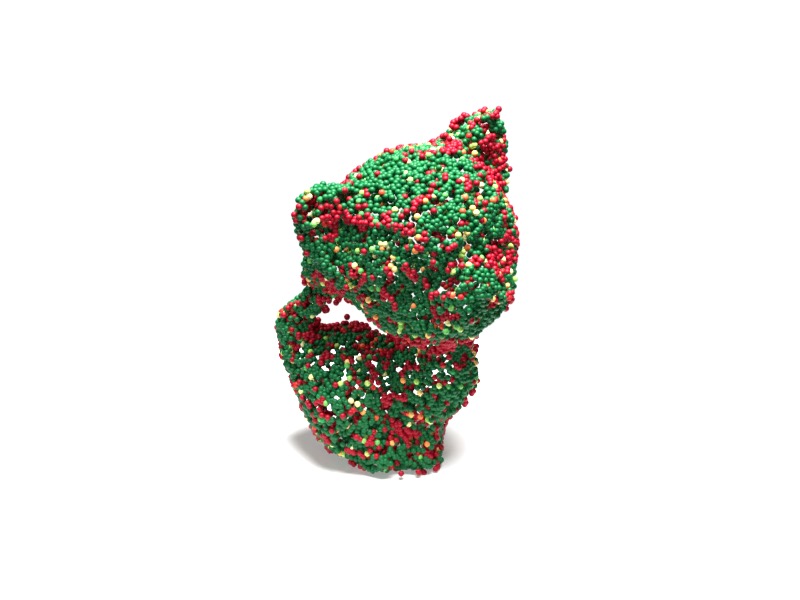}
\includegraphics[width=0.16\linewidth,trim={210 110 190 30},clip]{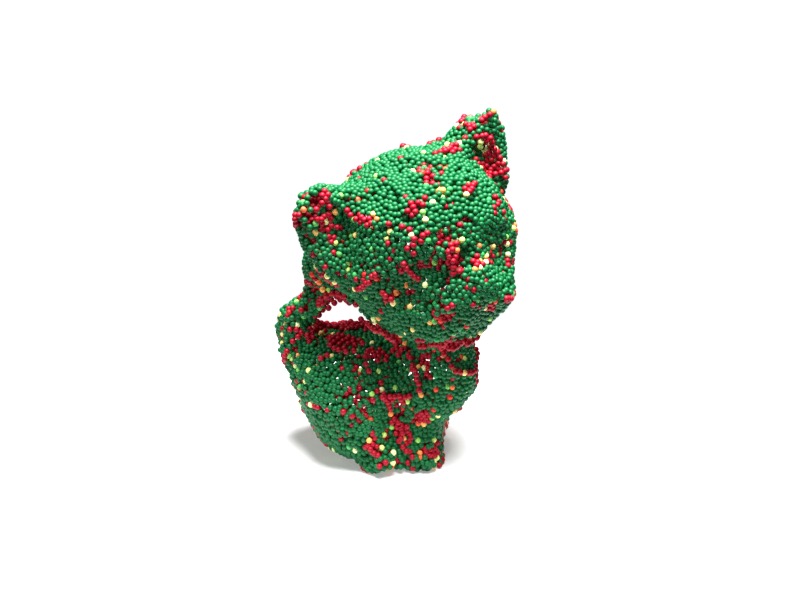}
\includegraphics[width=0.16\linewidth,trim={210 110 190 30},clip]{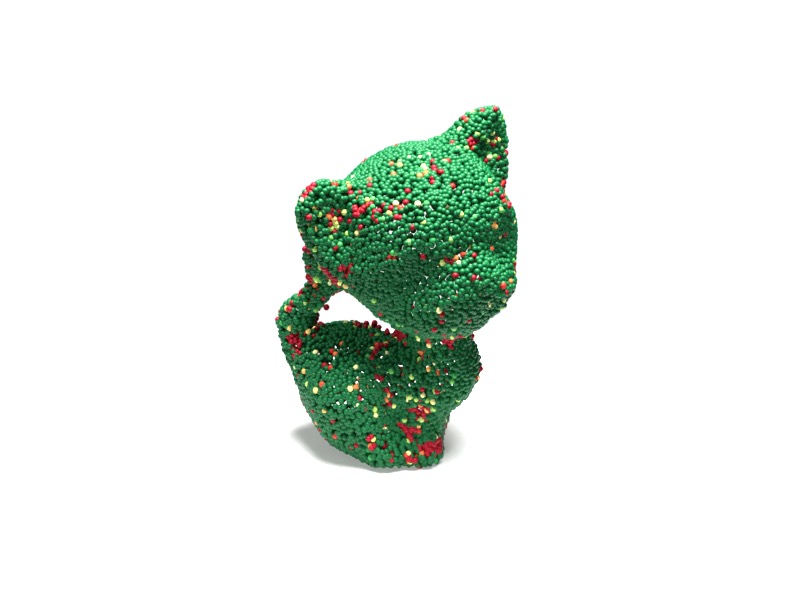}
\includegraphics[width=0.16\linewidth,trim={210 110 190 30},clip]{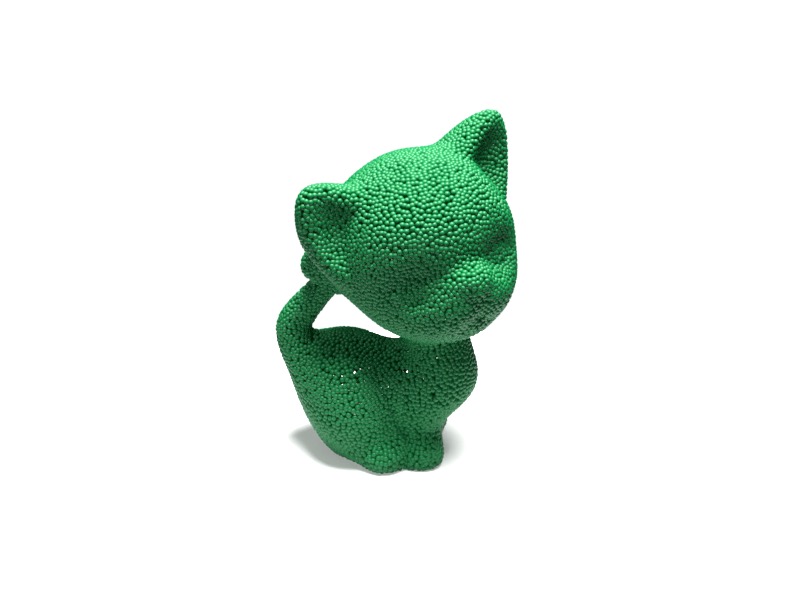}
\end{center}
\setlength{\tabcolsep}{0pt}
\vspace{-15px}
\resizebox{\textwidth}{!}{
\hspace{-10px}\begin{tabular}{cccccc}
\hspace{0.25\linewidth} &
\hspace{0.25\linewidth} &
\hspace{0.25\linewidth} &
\hspace{0.25\linewidth} &
\hspace{0.25\linewidth} &
\hspace{0.25\linewidth} \\
     \textbf{Noisy Input} & \textbf{ScoreDenoise}\cite{score_denoise} & \textbf{MAG}\cite{zhao2023point} & \textbf{PD-Flow}\cite{mao2022pd} & \textbf{\name{}} (Ours) & \textbf{Original} \\
\end{tabular}
}
\caption{
Qualitative comparison of our \name{} with recent deep-learning-based point cloud denoising methods on the PU-Net dataset under 3\% isotropic Gaussian noise.
}
\vspace{-22px}
\label{fig:pu_qualitative}
\end{figure}

\subsection{Comparison on Indoor Scenes}
To further investigate the denoising capabilities of our method and previous works,
we evaluate these methods on reconstructions of large-scale indoor scenes.
This setting introduces additional challenges and noise sources, such as clusters of outliers and surface thickening effects \cite{chen2018denoising_filter}, and highlights the methods' ability to scale to large inputs.

On ScanNet++, we evaluate all models on the noisy point cloud reconstructions provided by the authors.
These reconstructions were obtained by filtering the depth maps based on their agreement with the Faro laser depths, followed by projection using globally optimized poses without applying additional fusion methods \cite{yeshwanthliu2023scannetpp}.
Additionally, we evaluate all methods on reconstructions obtained by applying 3DMatch \cite{zeng20163dmatch} on the pre-filtered depth maps, using the globally optimized poses.
For ARKitScenes, we directly apply the methods to the ARKit reconstructions provided by the authors as part of the 3D object detection subset.
For further details about these reconstructions, we refer to the corresponding paper \cite{dehghan2021arkitscenes}.
\Cref{tab:evaluation_snpp} shows that our method, utilizing RGB+DINO features, generally achieves the best results, followed by our method using only RGB or coordinate features.
We qualitatively compare the best-performing methods in \cref{fig:snpp_arkit_qualitative}, including the noisy and Faro point clouds.
We use a color gradient to represent the distance between the predicted and ground truth points, ranging from green to red for low and high distances, respectively.
ScoreDenoise and other methods trained under the assumption of Gaussian noise exhibit pattern-like artifacts.
Due to memory constraints, all deep-learning methods denoise large point clouds in patches.
For methods trained under synthetic noise, this leads to a situation where points at the borders of each patch are treated as outliers, resulting in artificial borders around these patches. 
This effect is even more noticeable when Langevin sampling without stochasticity is used as in ScoreDenoise and MAG, leading to a collapse of points \cite{song2019generative}.
We hypothesize that our method is less susceptible to patch artifacts for two reasons.
First, we do not train under the assumption of Gaussian noise, making our method more robust in distinguishing real object borders from those arising due to patch-based processing.
Additionally, instead of simply accumulating predictions over patches, followed by farthest-point sampling, we average the predicted point coordinates for every point in the noisy cloud.
Although PD-Flow exhibits circular patch artifacts, it does not suffer from points collapsing on patch borders due to training on real-world noise.
Due to the lower level of detail in the noisy scans of ARKitScenes (\cf \cref{fig:snpp_arkit_qualitative}) compared to ScanNet++, the denoising is generally less pronounced.
Nonetheless, our method produces sharper object edges and smoother surfaces compared to other methods.
There are also incomplete objects visible, which none of the methods can fully reconstruct.
To address this, future work could incorporate strategies from point cloud completion \cite{lyu2022a, fei2022comprehensive} to further improve the results.

\begin{table}[ht]
\caption{\textbf{Denoising Results on Indoor Scenes.}
Quantitative point cloud denoising comparisons on scenes from the ScanNet++ \cite{yeshwanthliu2023scannetpp} and ARKitScenes \cite{dehghan2021arkitscenes} test set.
Quantitative point cloud denoising comparisons ScanNet++ \cite{yeshwanthliu2023scannetpp}, and ARKitScenes \cite{dehghan2021arkitscenes} test sets.
The Faro scanner point clouds serve as the reference for the ground truth. 
$\cdforw$ denotes the average distance from noisy points to the ground truth surface,
while $\cdbakw$ denotes the average distance from ground truth points to noisy points.
Similarly, P2F is the Point-to-Face distance, and F2P is the Face-to-Point distance.
On ScanNet++, scores are multiplied by $10^4$, on ARKitScenes by $10^3$.
}
	\centering
	\resizebox{\columnwidth}{!}{%
		\setlength{\tabcolsep}{3pt}
		\begin{tabular}{lccccccc|cccccc|ccc}
			\toprule
			& Dataset & \multicolumn{6}{c}{ScanNet++ \cite{yeshwanthliu2023scannetpp}} & \multicolumn{6}{c}{ScanNet++ \cite{yeshwanthliu2023scannetpp}} & \multicolumn{3}{c}{ARKitScenes \cite{dehghan2021arkitscenes}} \\
			& Input Source& \multicolumn{6}{c}{Apple LiDAR} & \multicolumn{6}{c}{Apple LiDAR + 3DMatch \cite{zeng20163dmatch}} & \multicolumn{3}{c}{Apple LiDAR}\\
			\cmidrule(lr){3-8} \cmidrule(lr){9-14} \cmidrule(lr){15-17}
			Method                            & Features       & P2F      & F2P        & $\cdforw$ & $\cdbakw$  & CD        & P2M       & P2F       & F2P       & $\cdforw$                       & $\cdbakw$ & CD        & P2M       & $\cdforw$ & $\cdbakw$ & CD        \\
			\midrule
			Bilateral \cite{digne_bilateral}  & XYZ            & 6.29     & 140.59     & 6.66      & 145.44     & 73.44     & 76.05     & 108.70    & \st 18.32 & 108.89                          & \st 19.67 & 64.28     & 63.51     & 15.87     & 70.49     & 43.18     \\
			DMR \cite{luo2020differentiable}  & XYZ            & 6.48     & 149.99     & 6.71      & 159.13     & 78.24     & 82.92     & 99.96     & 19.61     & 100.16                          & 21.27     & 60.71     & 59.79     & 10.84     & \rd 30.51 & 20.68     \\
			ScoreDenoise \cite{score_denoise} & XYZ            & 3.49     & \rd 128.59 & 3.72      & \rd 132.71 & 68.21     & 66.04     & 97.11     & 18.87     & 97.31                           & 20.26     & 58.78     & 57.99     & \nd 9.56  & 30.86     & 20.21     \\
			MAG \cite{zhao2023point}          & XYZ            & 5.43     & 147.54     & 5.66      & 152.07     & 78.87     & 76.49     & 99.05     & 24.82     & 99.26                           & 26.69     & 62.97     & 61.93     & \rd 9.57  & 30.82     & 20.20     \\
			PD-Flow \cite{mao2022pd}          & XYZ            & 3.80     & 147.49     & 4.02      & 151.90     & 77.96     & 75.64     & 85.29     & 21.00     & 85.49                           & 22.56     & 54.02     & 53.14     & 9.93      & 33.82     & 21.87     \\
			I-PFN \cite{de2023iterativepfn}   & XYZ            & 3.80     & 132.98     & 4.03      & 137.21     & 70.62     & 68.39     & 83.99     & 18.99     & 84.19                           & 20.43     & 52.31     & 51.49     & \st9.19   & 31.99     & 20.59     \\
			\textbf{\name{} (Ours)}           & XYZ            & \rd 2.48 & \st 122.23 & \rd 2.71  & \st 126.22 & \nd 64.46 & \nd 62.35 & \rd 50.87 & 18.69     & \rd 51.07                       & 20.05     & \rd 35.56 & \rd 34.78 & 9.65      & 30.64     & \rd 20.14 \\
			\midrule
			\textbf{\name{} (Ours)}           & XYZ, RGB       & \nd 2.47 & \nd 122.27 & \nd 2.70  & \nd 126.26 & \rd 64.48 & \rd 62.37 & \nd 50.40 & \nd 18.39 & \nd 50.60                       & \nd 19.73 & \nd 35.17 & \nd 34.39 & 9.65      & \nd 30.45 & \nd 20.05 \\
			\textbf{\name{} (Ours)}           & XYZ, RGB, DINO & \st 2.42 & \st 122.23 & \st 2.65  & \st 126.22 & \st 64.44 & \st 62.33 & \st 49.64 & \rd 18.57 & \st 49.84                       & \rd 19.92 & \st 34.88 & \st 34.11 & \rd 9.57  & \st 30.27 & \st 19.92 \\
			\bottomrule
		\end{tabular}
	}
	\label{tab:evaluation_snpp}
 \vspace{-10px}
\end{table}

\begin{figure}[htb!]
\begin{center}
\vspace{12px}
    \resizebox{0.5\textwidth}{!}{
    \begin{tabular}{rcl}
Noisy &\includegraphics[width=0.6\linewidth, height=8px]{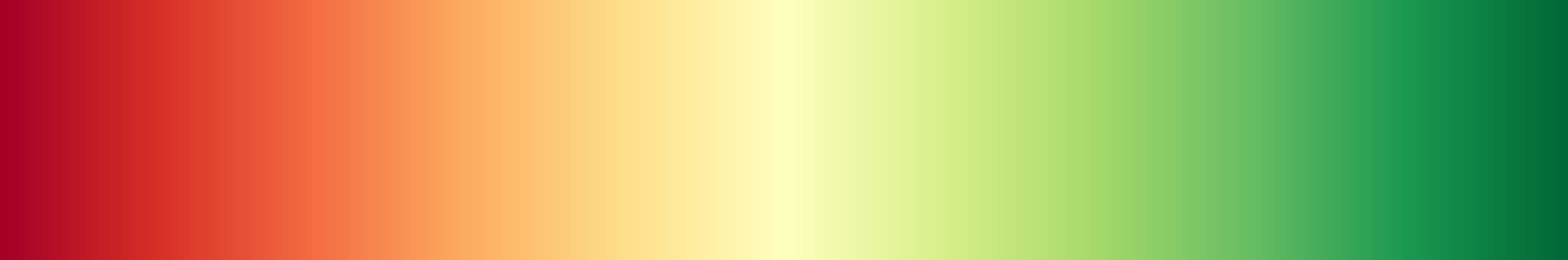} & Clean
    \end{tabular}
}
\end{center}

\newcommand{\tabspacesnpp}{0.16}

\begin{center}

\includegraphics[width=\tabspacesnpp\linewidth,clip]{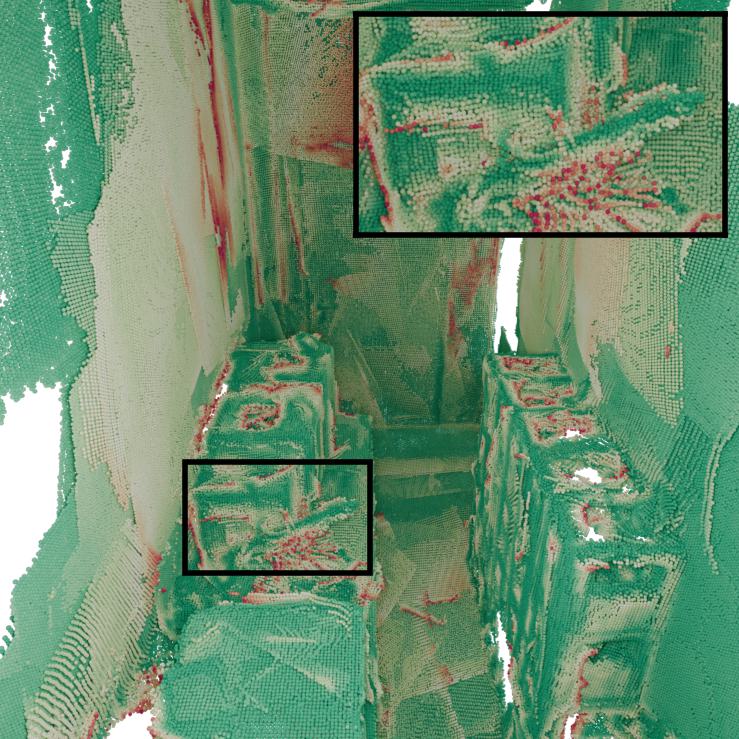}
\includegraphics[width=\tabspacesnpp\linewidth,clip]{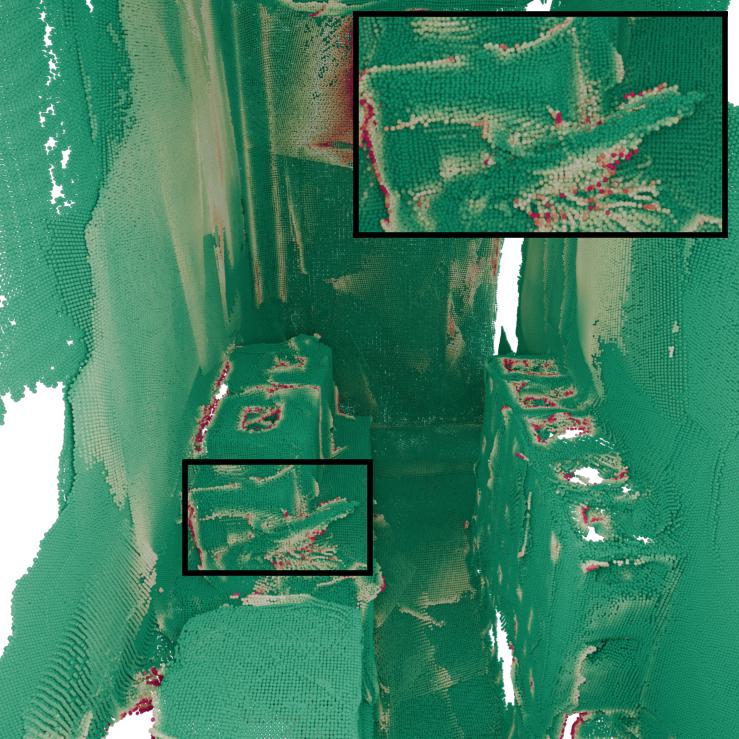}
\includegraphics[width=\tabspacesnpp\linewidth,clip]{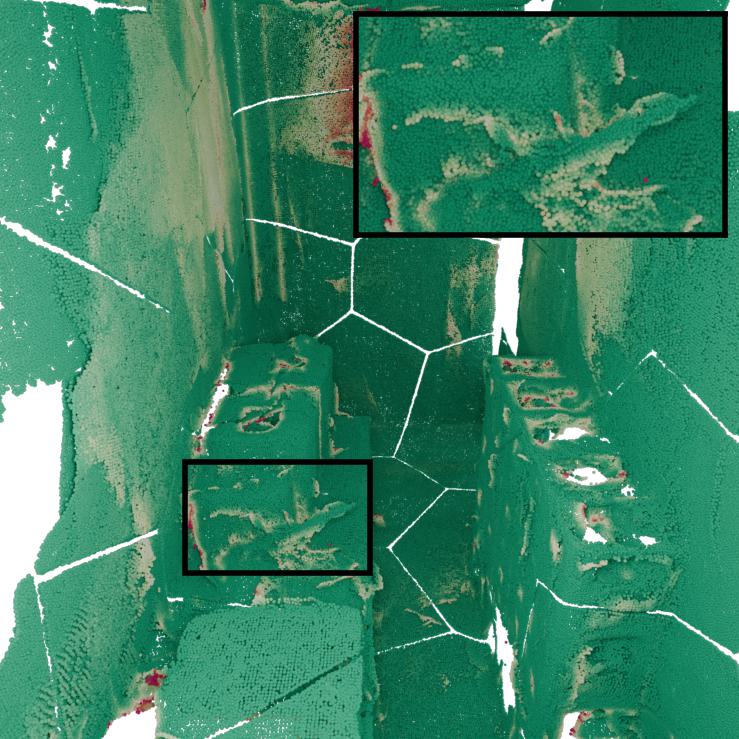}
\includegraphics[width=\tabspacesnpp\linewidth,clip]{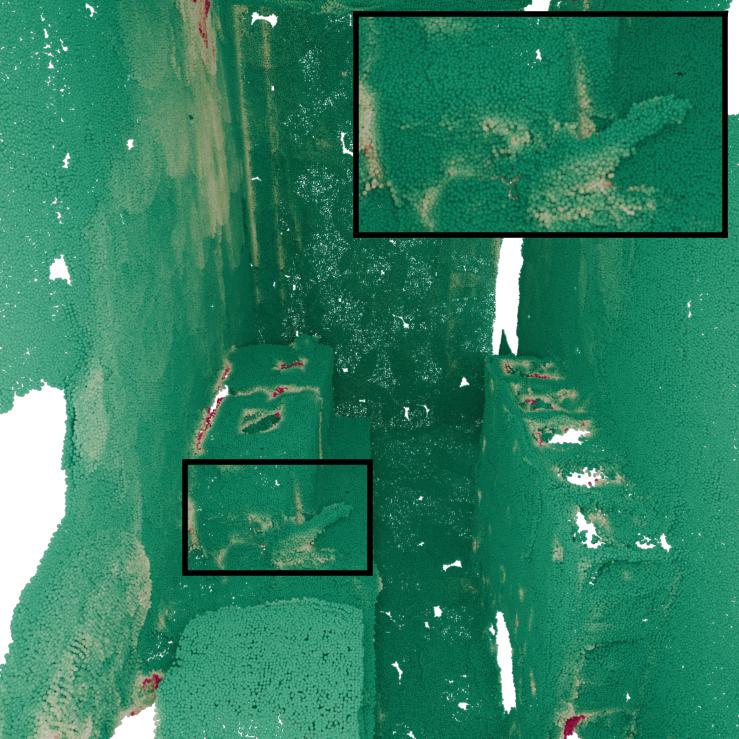}
\includegraphics[width=\tabspacesnpp\linewidth,clip]{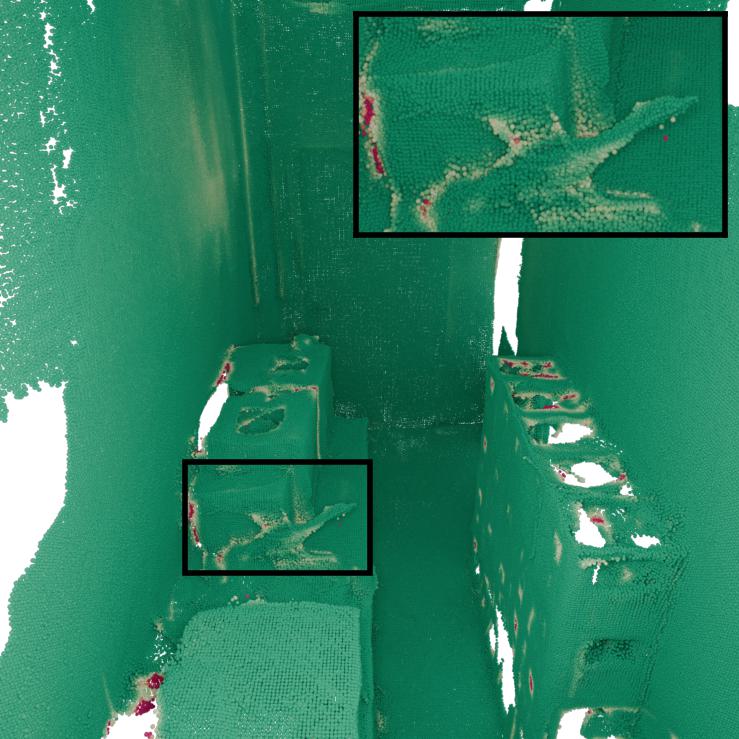}
\includegraphics[width=\tabspacesnpp\linewidth,clip]{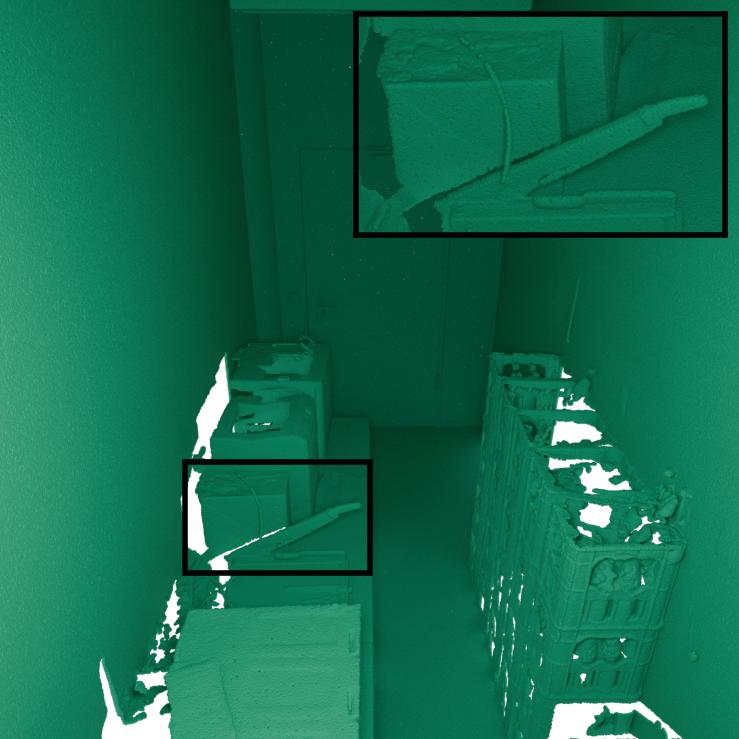}
\\
\includegraphics[width=\tabspacesnpp\linewidth,clip]{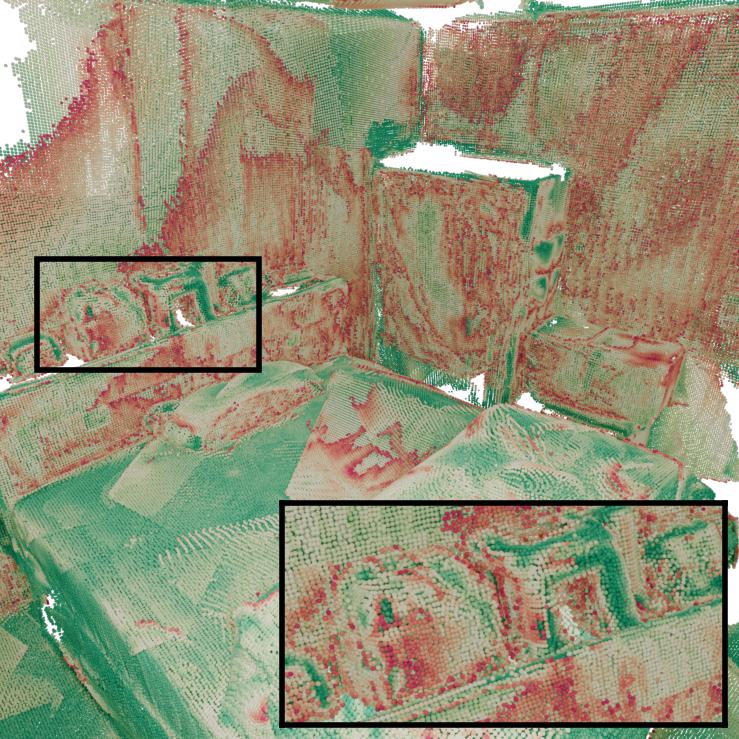}
\includegraphics[width=\tabspacesnpp\linewidth,clip]{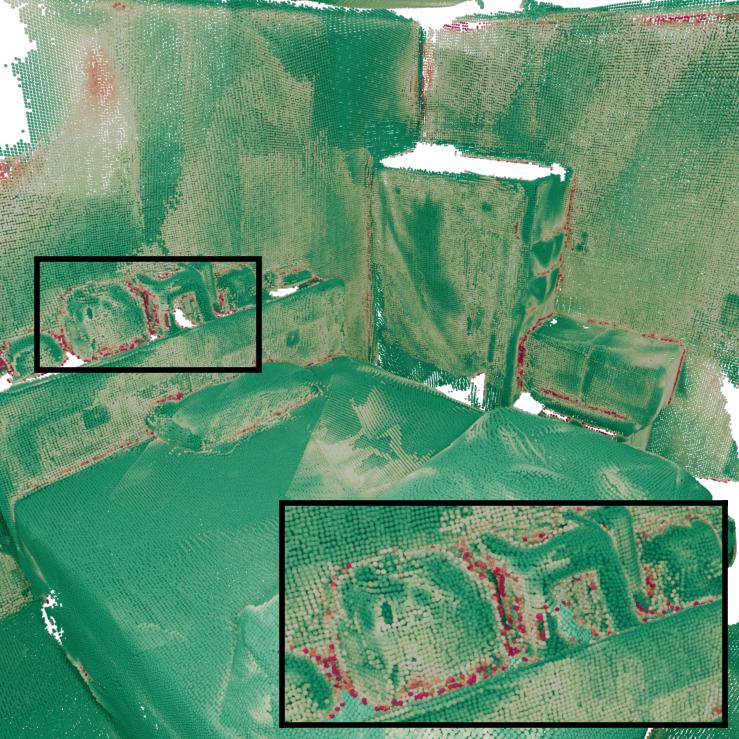}
\includegraphics[width=\tabspacesnpp\linewidth,clip]{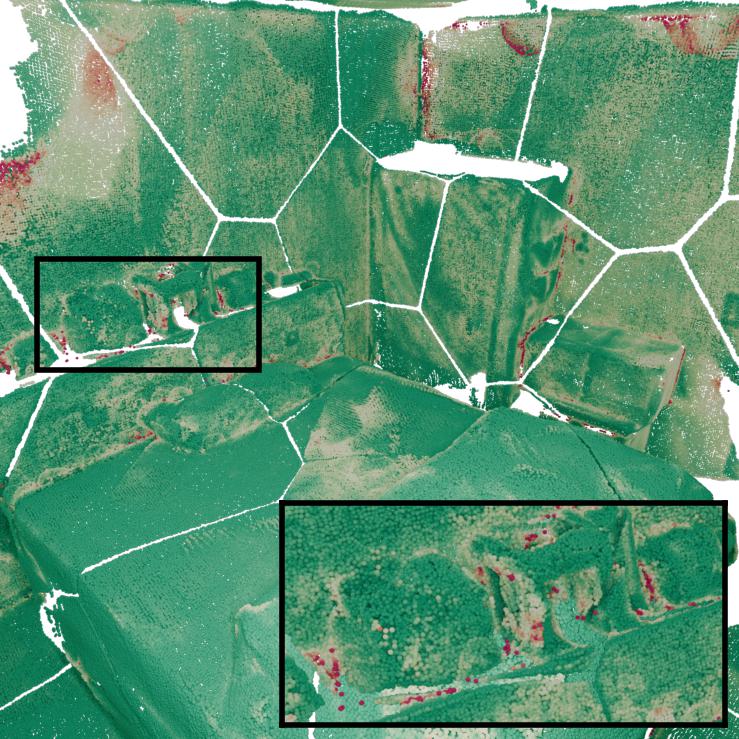}
\includegraphics[width=\tabspacesnpp\linewidth,clip]{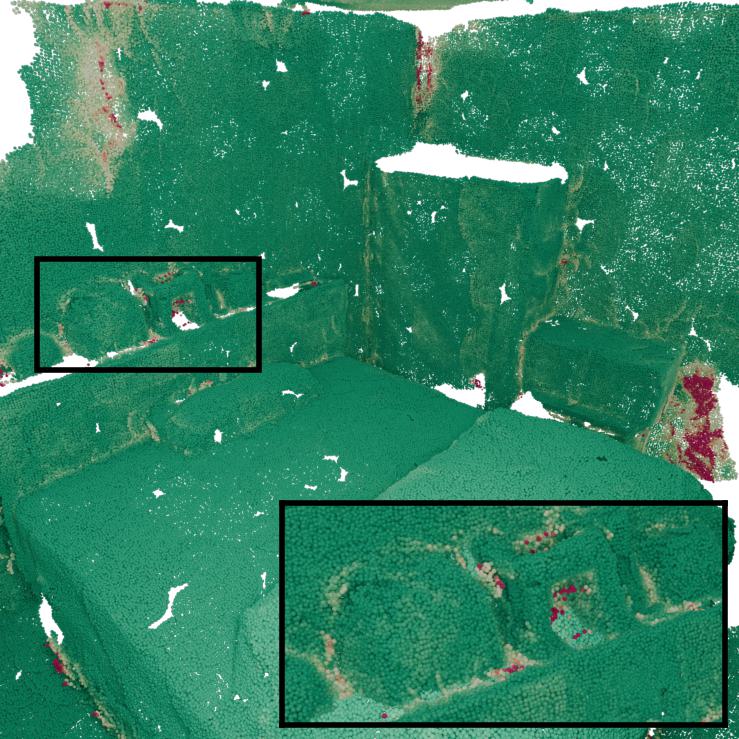}
\includegraphics[width=\tabspacesnpp\linewidth,clip]{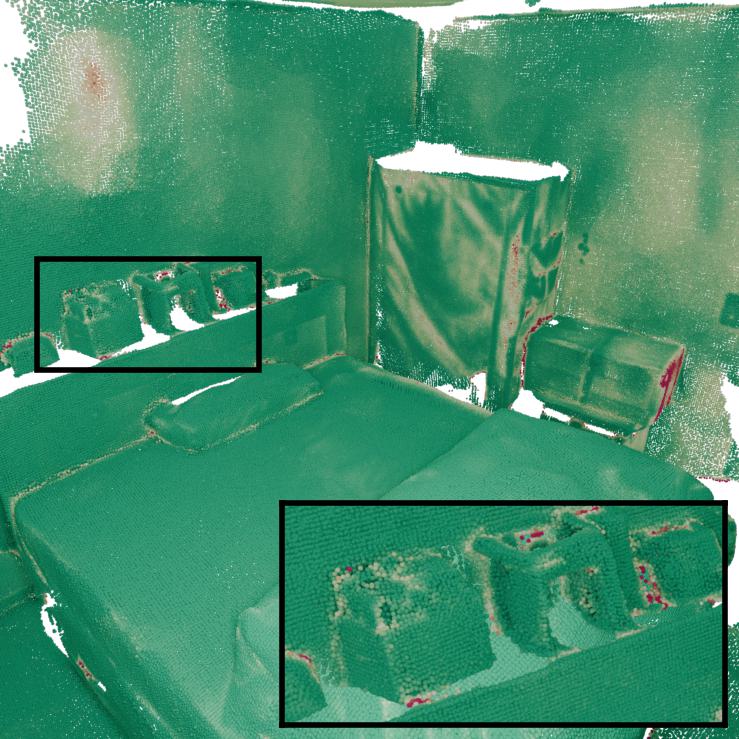}
\includegraphics[width=\tabspacesnpp\linewidth,clip]{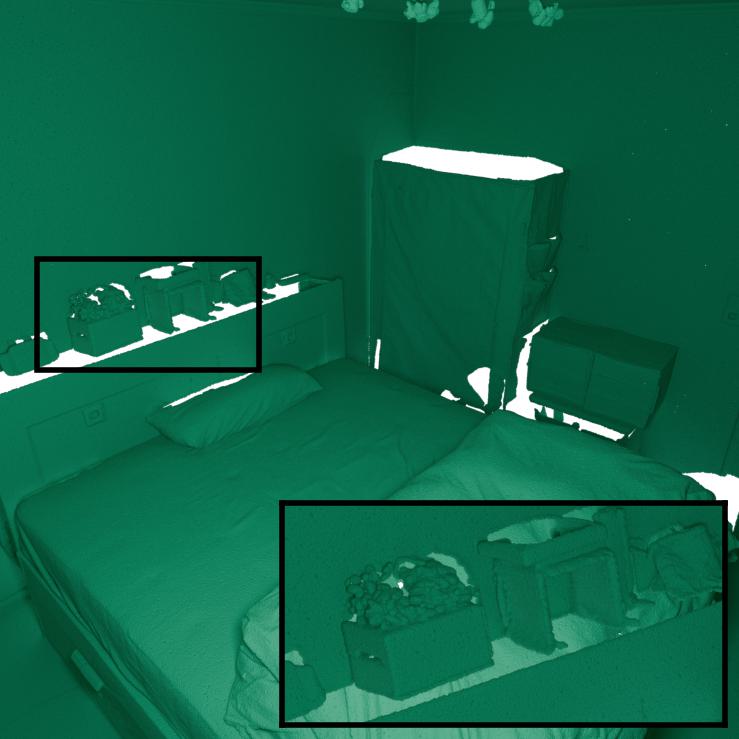}
\\
\includegraphics[width=\tabspacesnpp\linewidth,clip]{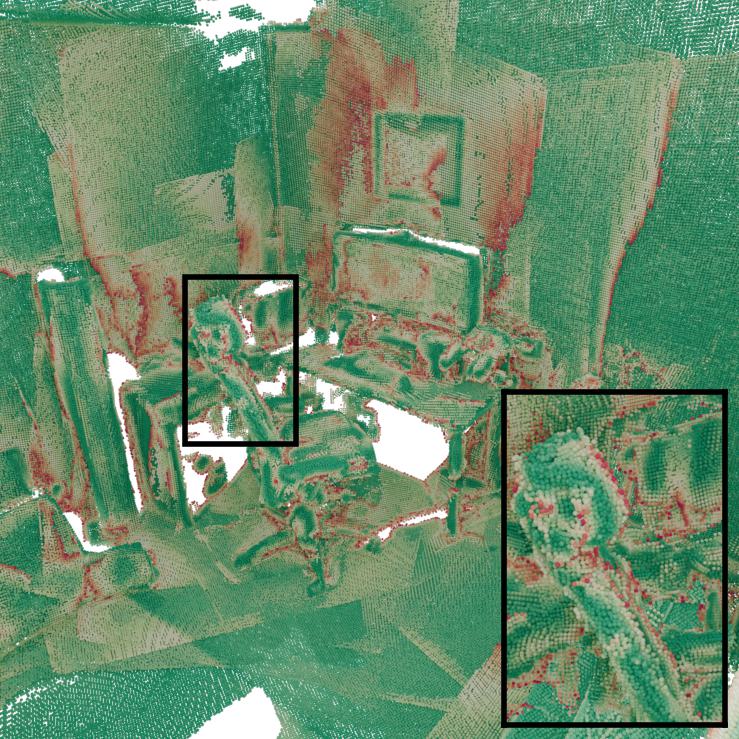}
\includegraphics[width=\tabspacesnpp\linewidth,clip]{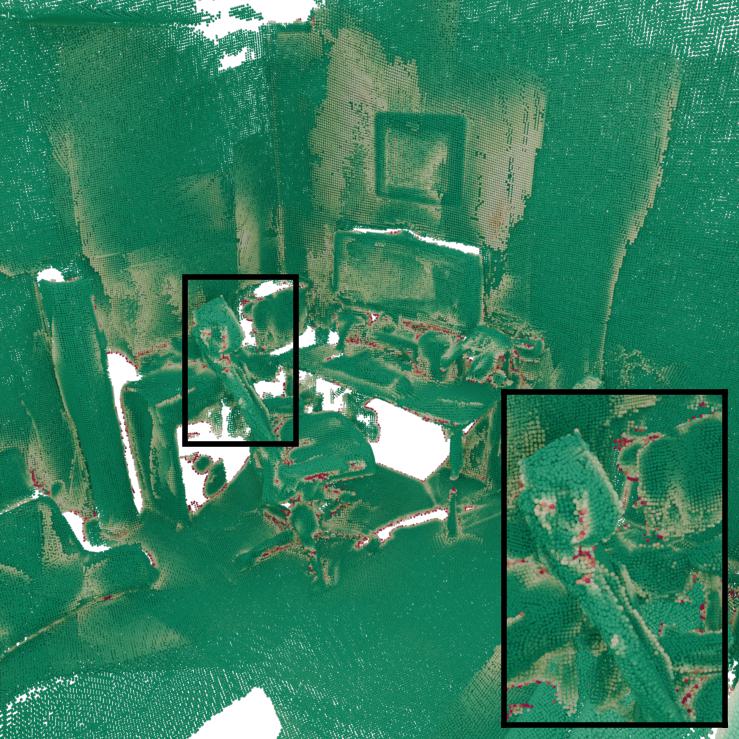}
\includegraphics[width=\tabspacesnpp\linewidth,clip]{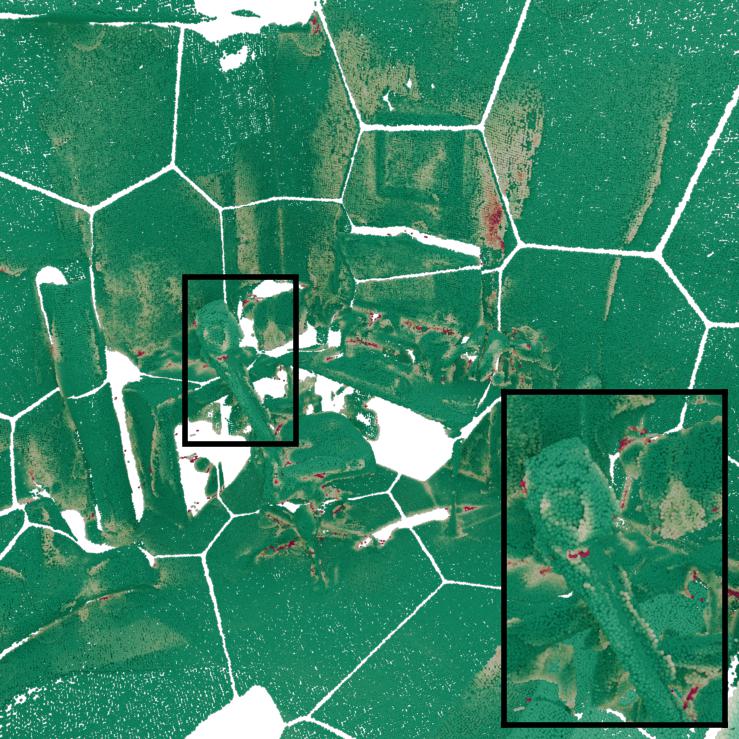}
\includegraphics[width=\tabspacesnpp\linewidth,clip]{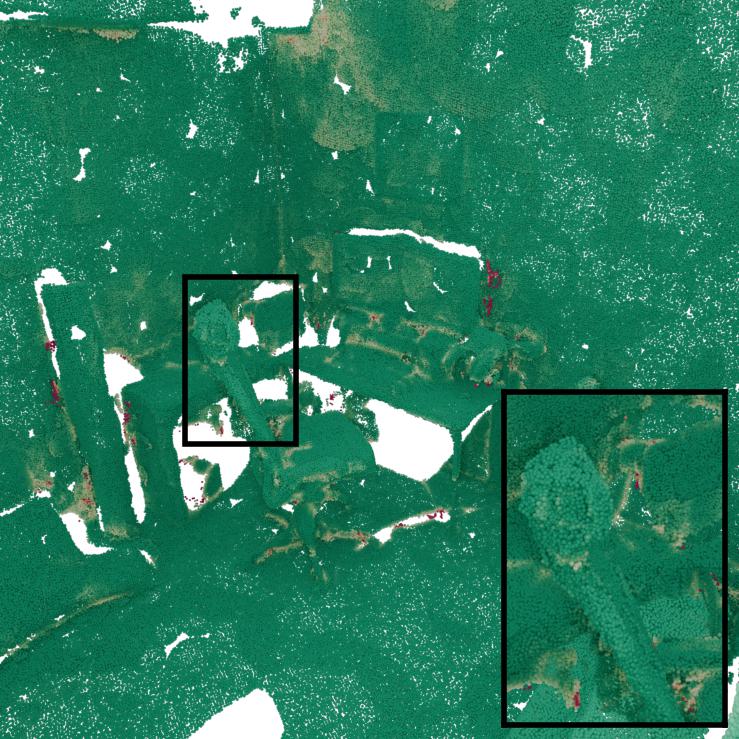}
\includegraphics[width=\tabspacesnpp\linewidth,clip]{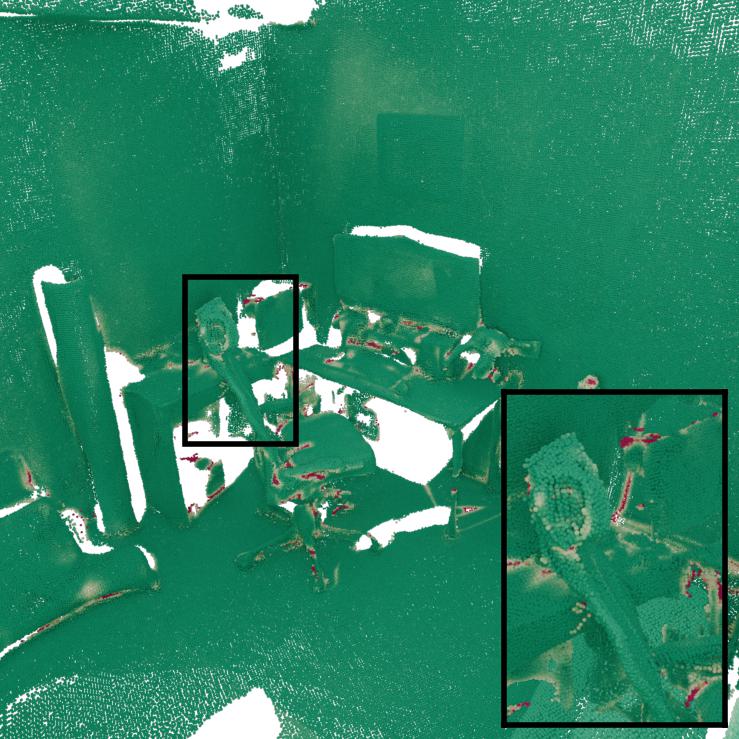}
\includegraphics[width=\tabspacesnpp\linewidth,clip]{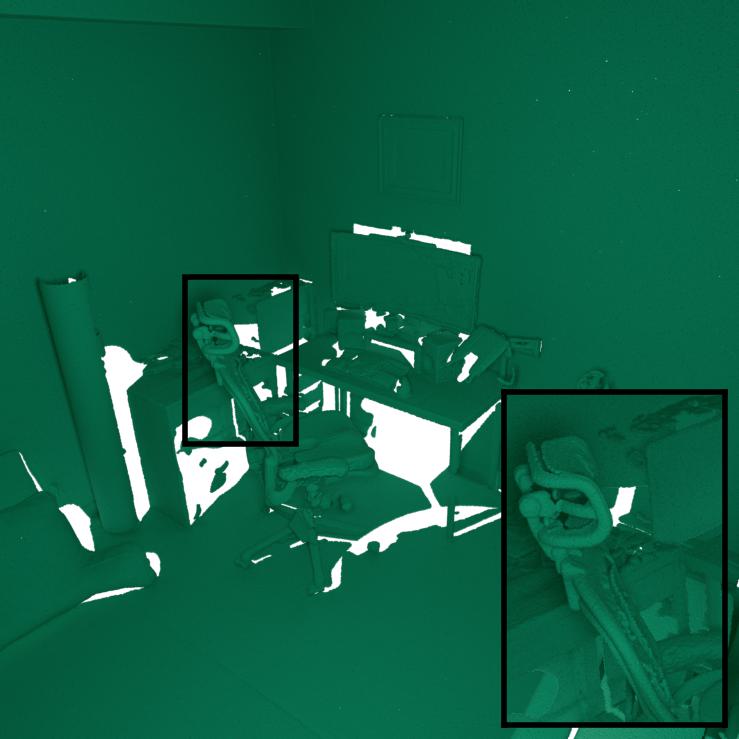}
\\
\includegraphics[width=\tabspacesnpp\linewidth,clip]{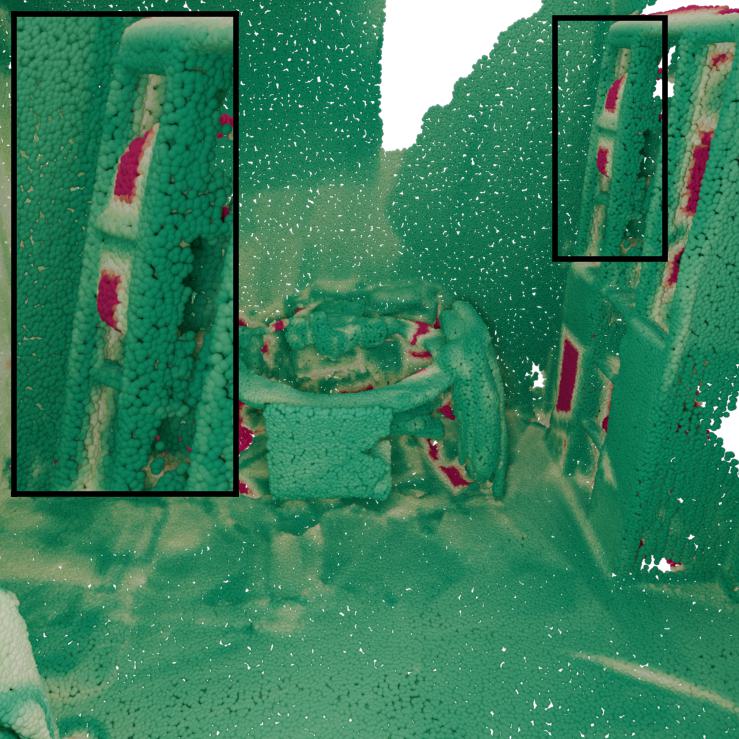}
\includegraphics[width=\tabspacesnpp\linewidth,clip]{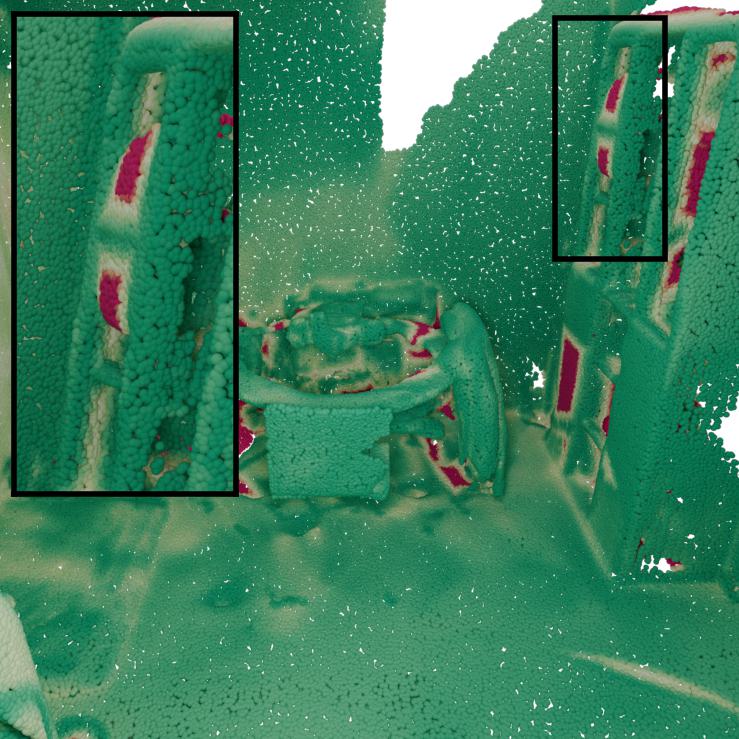}
\includegraphics[width=\tabspacesnpp\linewidth,clip]{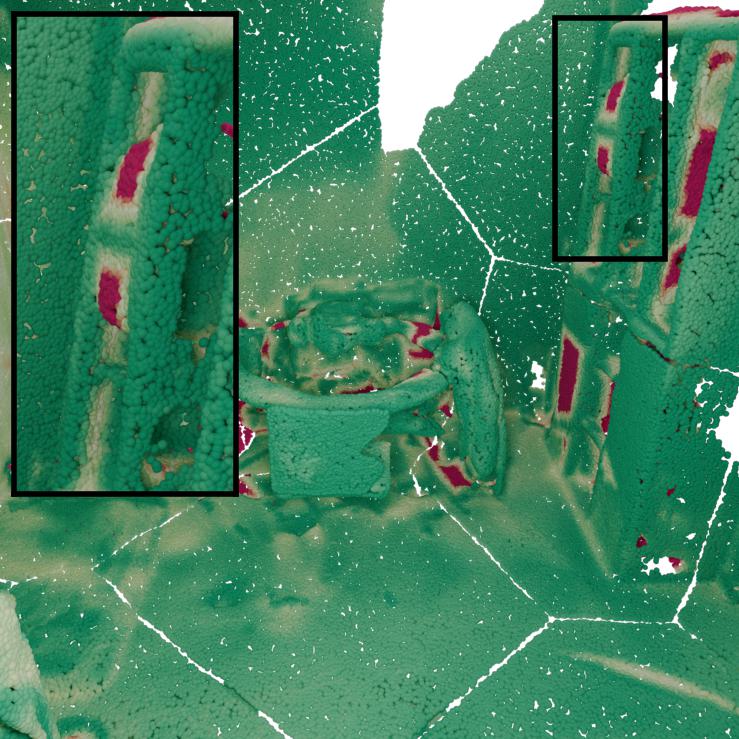}
\includegraphics[width=\tabspacesnpp\linewidth,clip]{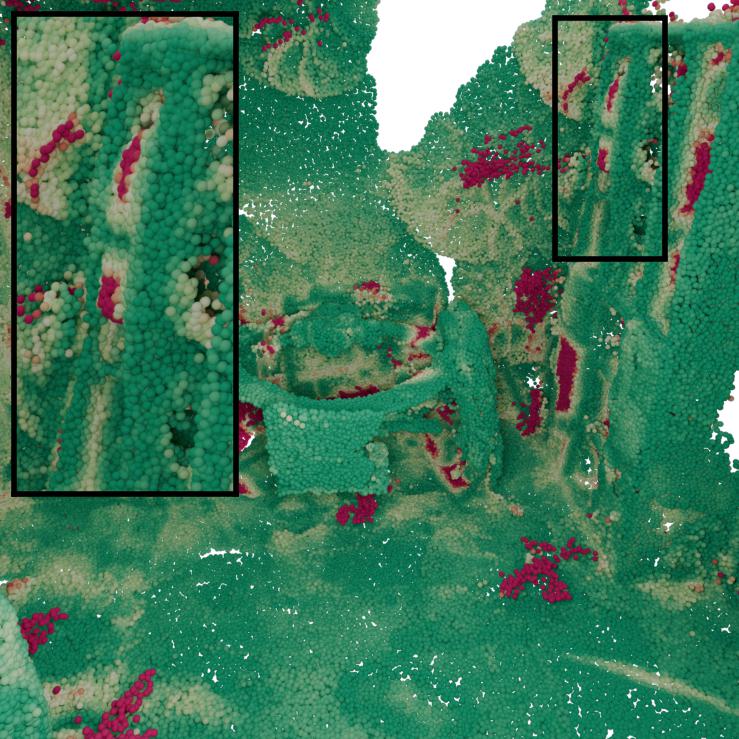}
\includegraphics[width=\tabspacesnpp\linewidth,clip]{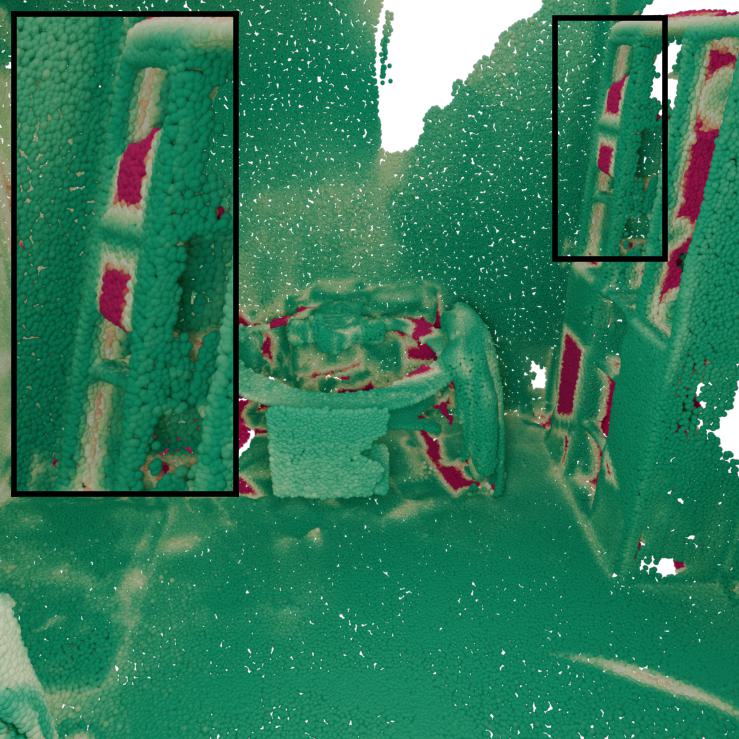}
\includegraphics[width=\tabspacesnpp\linewidth,clip]{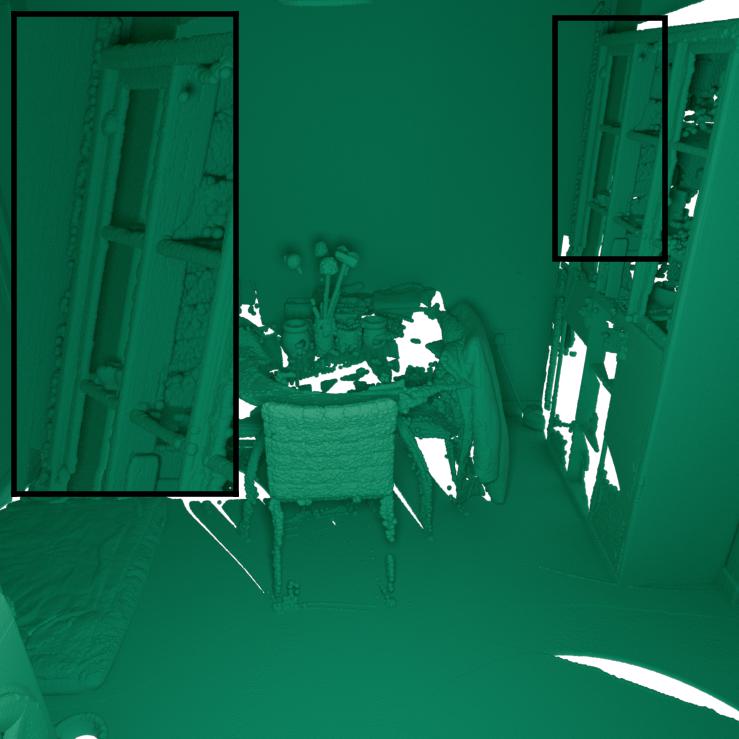}
\\
\includegraphics[width=\tabspacesnpp\linewidth,clip]{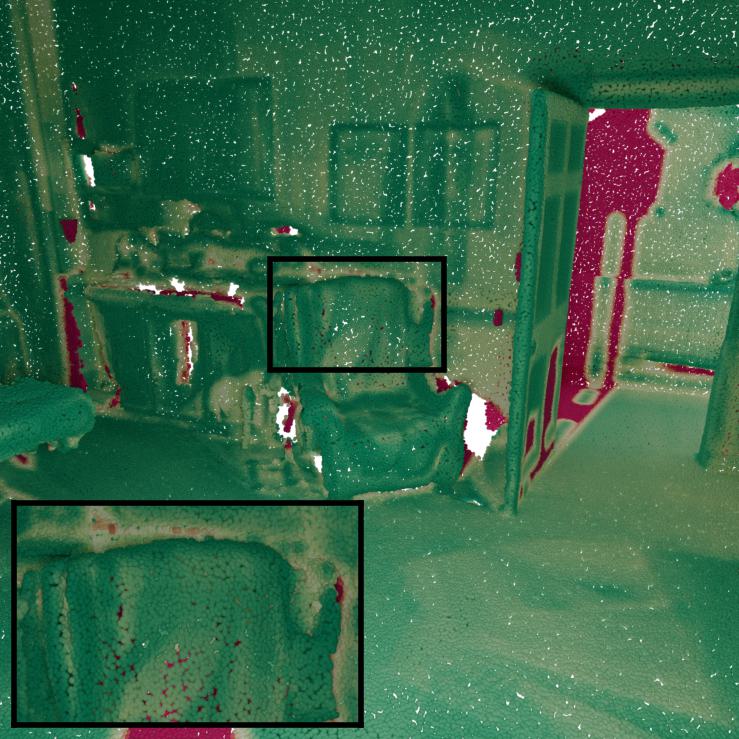}
\includegraphics[width=\tabspacesnpp\linewidth,clip]{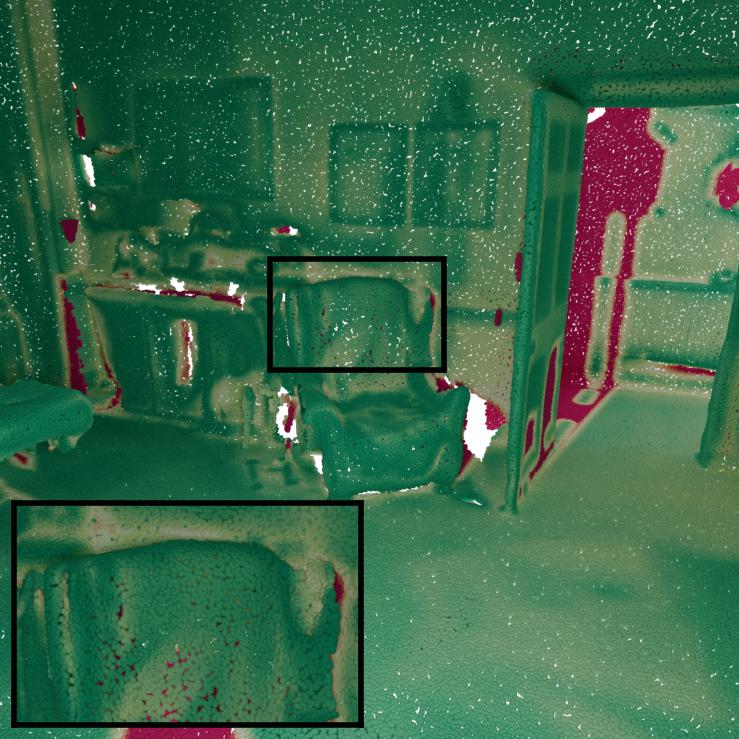}
\includegraphics[width=\tabspacesnpp\linewidth,clip]{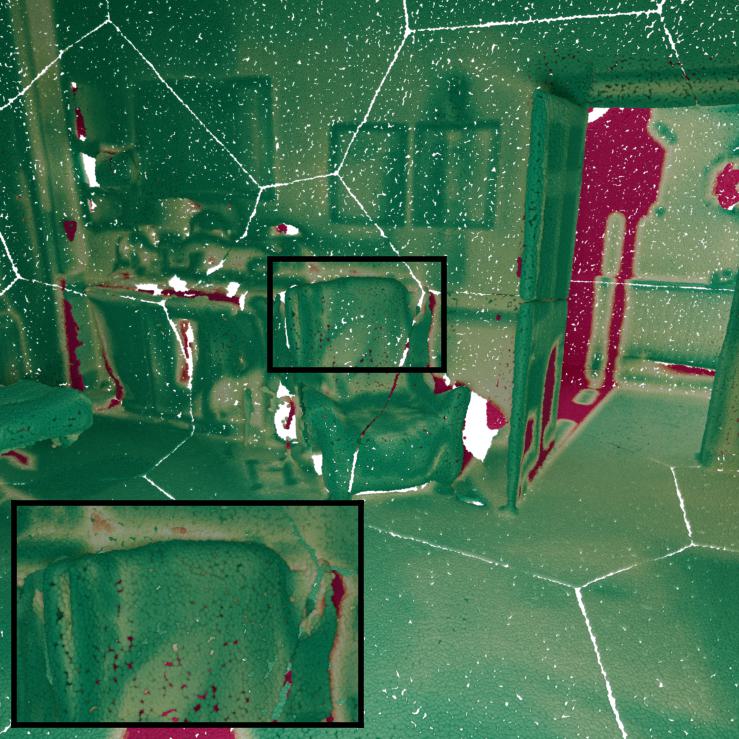}
\includegraphics[width=\tabspacesnpp\linewidth,clip]{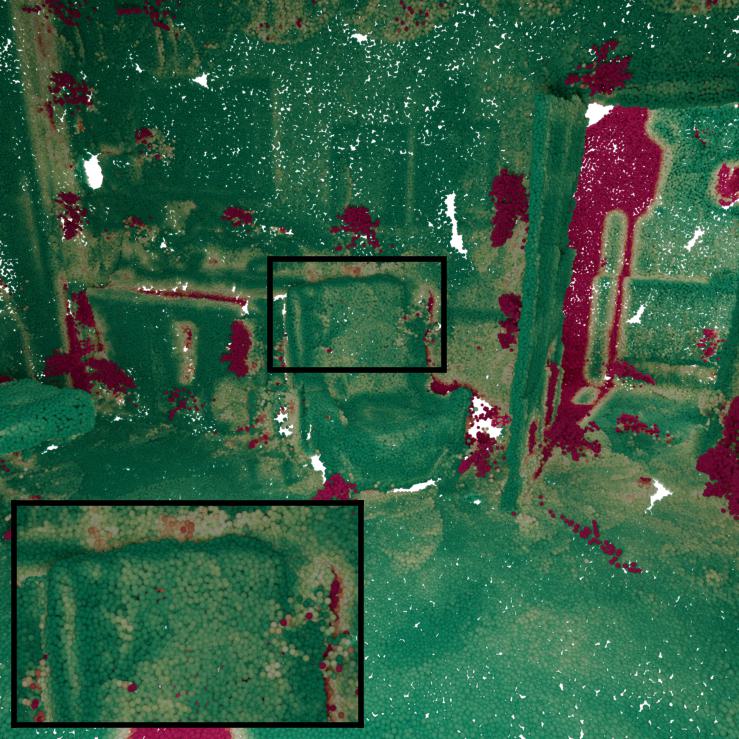}
\includegraphics[width=\tabspacesnpp\linewidth,clip]{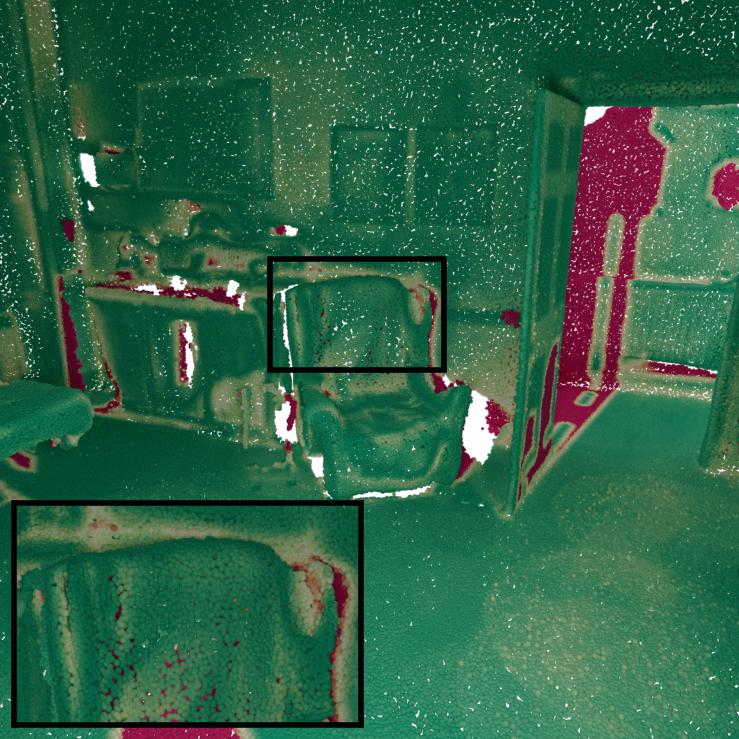}
\includegraphics[width=\tabspacesnpp\linewidth,clip]{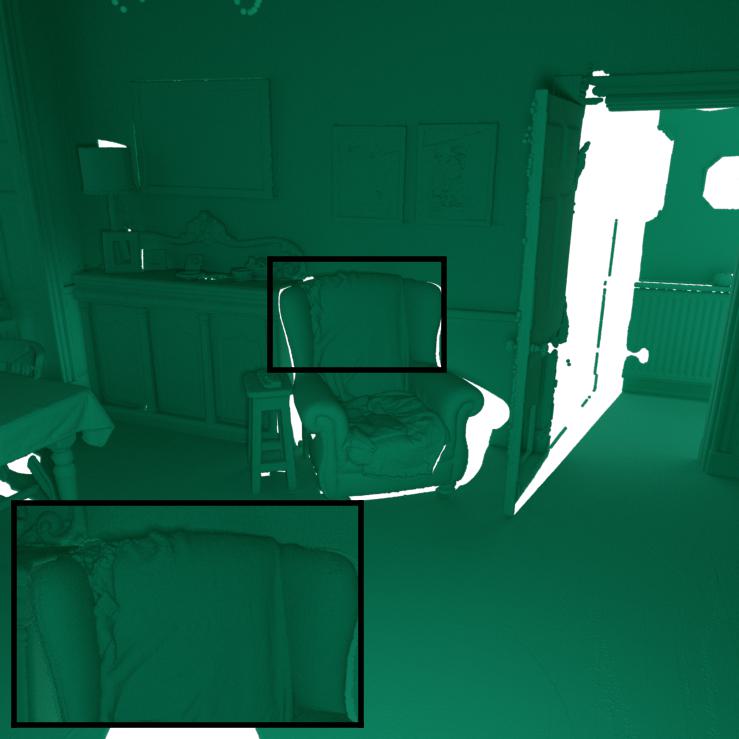}

\end{center}
\setlength{\tabcolsep}{2pt}
\vspace{-15px}
\resizebox{\textwidth}{!}{
\hspace{0px}\begin{tabular}{cccccc}
\hspace{0.25\linewidth} &
\hspace{0.25\linewidth} &
\hspace{0.25\linewidth} &
\hspace{0.25\linewidth} &
\hspace{0.25\linewidth} &
\hspace{0.25\linewidth} \\
     \textbf{Noisy Input} & \textbf{Bilateral}\cite{digne_bilateral} &\textbf{ScoreDenoise}\cite{score_denoise} & \textbf{PD-Flow}\cite{mao2022pd} & \textbf{\name{}} & \textbf{Ground Truth} \\
     &&&& \textbf{(Ours)} &\textbf{(Faro)} \\ 
\end{tabular}
}
\caption{
Qualitative comparison on the ScanNet++ dataset \cite{yeshwanthliu2023scannetpp} (top 3 rows)
and the ARKitScenes dataset \cite{dehghan2021arkitscenes} (bottom 2 rows)
using noisy iPhone scans as input.}
\label{fig:snpp_arkit_qualitative}
\end{figure}

\subsection{Ablation Studies}

\parag{Diffusion Model Backbones.}
We evaluate various diffusion model backbones on a subset of the ScanNet++ dataset.
Specifically, we consider an architecture based on the Point-Voxel-Convolutional Neural Network (PVCNN) \cite{pvcnn} (\cf \cref{fig:architecture}), a transformer-based architecture from GECCO \cite{tyszkiewicz2023gecco}, and a sparse-convolution-based architecture based on the Minkowski engine \cite{minkowski}.
Since the Minkowski Engine does not provide pre-defined skeletons for diffusion model architectures, we recreate the DDPM \cite{ddpm} backbone architecture using only building blocks from Minkowski.
\Cref{tab:backbone_analysis} presents the resulting performance, with the PVCNN architecture showing the best results.
\parag{Bridge Settings.}
We evaluate the impact of nearest-neighbor interpolation and stochasticity on PU-Net.
\Cref{tab:bridge_ablation} shows that training without prior alignment of the unordered point cloud data significantly reduces the performance of our method.
Indeed, without proper data alignment, the method fails to converge.
Introducing stochasticity into the interpolation path during training, effectively training an SDE instead of an OT-ODE, also degrades performance.
We speculate that this is due to the strong prior information embedded in noisy scans.
However, for tasks such as point cloud completion, stochasticity may be necessary \cite{lyu2022a, i2sb}.

\noindent
\begin{minipage}[ht!]{.49\linewidth}
    \centering
    \captionof{table}{Bridge settings comparison on the PU-Net dataset.
    CD and P2M are both multiplied by $10^4$.}
    \resizebox{.7\linewidth}{!}{%
    \begin{tabular}{cccc}
        \toprule
         OT-ODE & Alignment &  CD & P2M\\
         \midrule
         \cmark &  \xmark & \rd 49.33 & \rd 44.22\\
         \xmark & \cmark & \nd 2.45 & \nd 0.73\\
         \cmark &  \cmark & \st 2.11  & \st 0.65\\
         \bottomrule
    \end{tabular}}
   \label{tab:bridge_ablation}
\end{minipage}%
\hfill
\begin{minipage}[ht!]{.49\linewidth}
    \centering
    \captionof{table}{Diffusion model backbones comparison on the ScanNet++ dataset. CD and P2M are both multiplied by $10^4$.}
    \label{tab:backbone_analysis} 
    \resizebox{.7\linewidth}{!}{%
    \begin{tabular}{lcc}
        \toprule
         Backbone & CD & P2M\\
         \midrule
         Minkowksi \cite{minkowksi_4d} & \rd 12.79 & \rd 32.39\\
         SetTransformer \cite{tyszkiewicz2023gecco} & \nd 10.50 & \nd 15.46\\
         PVCNN \cite{pvcnn} & \st 9.76 & \st 14.33\\
         \bottomrule
    \end{tabular}}
\end{minipage}%

\begin{figure*}[t]
\begin{minipage}[b]{0.58\textwidth}
    \centering
    \captionof{table}{
    Model analysis on ScanNet++.
    Squeeze-and-excitation (SE) blocks \cite{hu2018squeeze} are applied after convolutional layers.
   Chamfer Distance (CD) and Point-to-Mesh (P2M) scores are multiplied by $10^4$. \vspace{0px}}
    \resizebox{\textwidth}{!}{%
    \setlength{\tabcolsep}{3pt}
    \begin{tabular}{ccccccc}
        \toprule
         Base & PVC & Global &  & Feature & & \\
         Channels & Blocks & Feature & SE & Embedding & CD ($\Delta$)& P2M ($\Delta$)\\
         \midrule
         32 & 1222 & \cmark & \cmark & \cmark & 9.41 $(+1.14)$& 13.78 $(+4.08)$\\
         32 & 2122 & \cmark & \cmark & \cmark & 9.45 $(+1.18)$& 13.82 $(+4.12)$\\
         32 & 2212 & \cmark & \cmark & \cmark & 9.33 $(+1.06)$& 13.70 $(+4.00)$\\
         32 & 2221 & \cmark & \cmark & \cmark & 9.32 $(+1.05)$& 13.70 $(+4.00)$\\
         32 & 2222 & \cmark & \cmark & \cmark & 9.26 $(+0.99)$& 13.67 $(+3.97$\\
         \midrule
         32 & 2222 & \xmark & \cmark & \cmark & 9.36 $(+1.09)$& 13.75 $(+4.05)$\\
         32 & 2222 & \cmark & \xmark & \cmark & 9.33 $(+1.06)$& 13.71 $(+4.01)$\\
         32 & 2222 & \cmark & \cmark & \xmark & 9.76 $(+1.49)$& 14.33 $(+4.61)$\\
         \midrule 
         64 & 2222 & \cmark & \cmark & \cmark & 8.31 $(+0.04)$& 9.72 $(+0.02)$\\
         64 & 2322 & \cmark & \cmark & \cmark & $\mathbf{8.27~(--)}$ & $\mathbf{9.70~(--)}$\\
         \bottomrule
    \end{tabular}%
    }
    \label{tab:network_ablation}
\end{minipage}
\hfill
\begin{minipage}[b]{0.38\textwidth}
    \centering
    \includegraphics[width=\textwidth]{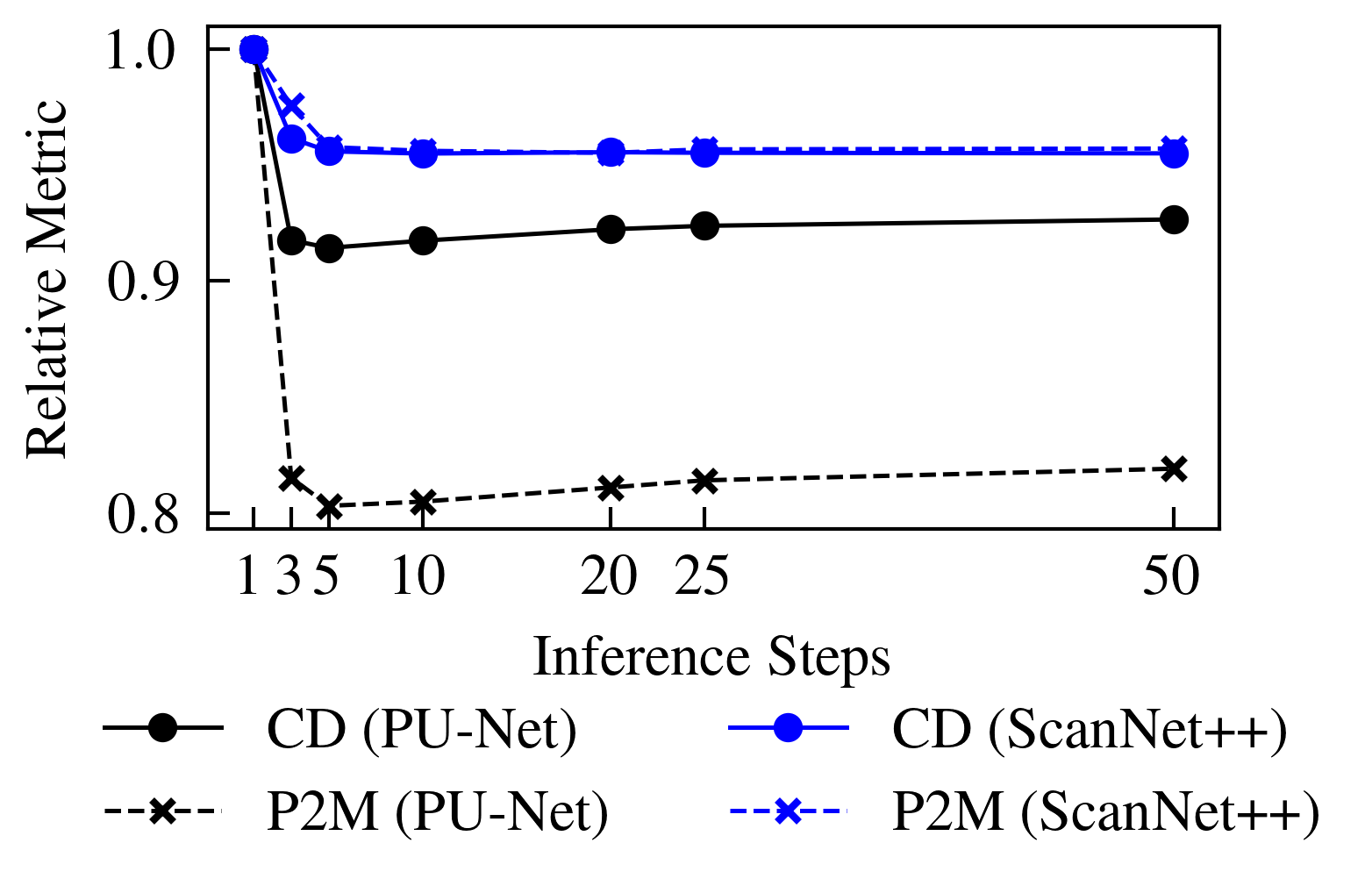}
    \captionof{figure}{
    Relative improvement in Chamfer Distance (CD) and Point-to-Mesh (P2M) with increasing sampling steps.
    Good scores are achieved with as few as 5 steps, enabling fast inference.}
    \label{fig:inference-steps}
\end{minipage}
\end{figure*}

\parag{Model Analysis.}
We investigate the importance of individual building blocks and their attributes in \cref{tab:network_ablation}.
The study indicates that increasing the number of blocks generally improves results, with a more pronounced difference in shallower blocks.
Since the input is down-sampled after each SA-Block, shallower blocks can extract more fine-grained features, which may explain their larger impact, as the voxel convolutions and the global feature network already capture coarse features.
Among additional feature layers, the feature embedding has the most significant impact.  
However, doubling the channels in each layer results in the greatest improvement in evaluation metrics.
\Cref{fig:inference-steps} illustrates the relative change in metrics with increasing inference steps.
Good results can be achieved with as few as five to ten inference steps, after which the metrics tend to plateau.

\noindent

\section{Conclusion and Discussion}
In this paper, we introduced \name{}, a point cloud denoising framework based on diffusion Schrödinger bridges.
Our approach addresses the denoising task as a data-to-data diffusion problem by learning an optimal transport path between point sets.
We emphasized the importance of data alignment when applying diffusion bridges to point clouds, drawing parallels between optimal transport plans and shortest-path point cloud interpolation, and empirically demonstrated the effectiveness of this approach.
Our method was tested on both single object datasets and large-scale indoor point clouds, and extensive experiments showed that it outperforms previous methods in both contexts.
Additionally, we demonstrated that incorporating image-based features, such as color information, and point-wise high-level features, such as DINOv2 features, further enhances the point cloud denoising performance.

\parag{Acknowledgements} Francis Engelmann is partially supported by a postdoctoral research fellowship from the ETH AI Center and an ETHZ Career Seed Award.

%
%
\bibliographystyle{splncs04}
\bibliography{main}

\end{document}